\documentclass[10pt,twocolumn,letterpaper]{article}

\usepackage[pagenumbers]{cvpr} % To force page numbers, e.g. for an arXiv version

\usepackage{nicematrix}
\usepackage{sg-macros}
\usepackage{pifont}
\usepackage{multirow}
\usepackage{xcolor} % Required for defining colors
\usepackage{soul}   % For the \hl command
\usepackage{subcaption}
\usepackage{graphicx}
\usepackage{makecell} 
\usepackage[table]{xcolor}

\newcommand{\OurDatasetName}{RoMo}
\renewcommand{\subsubsection}[1]{\noindent\textbf{#1}}

\definecolor{cvprblue}{rgb}{0.21,0.49,0.74}
\usepackage[pagebackref,breaklinks,colorlinks,allcolors=cvprblue]{hyperref}

\def\paperID{10194}
\def\confName{CVPR}
\def\confYear{2026}

\title{
RoMo: A Large-Scale, Richly Organized Dataset and Semantic Taxonomy for Human Motion Generation
}

\author{
Jiahao Zhang$^{1,2}$\quad
Joseph Liu$^{2}$\quad
Young-Yoon Lee$^{2}$\quad
Seonghyeon Moon$^{2}$\\
Victor Zordan$^{2}$\quad
Guy Tevet$^{3}$\quad
C. Karen Liu$^{3}$\quad
Stephen Gould$^{1}$\\
Oren Jacob$^{2}$\quad
Haomiao Jiang$^{2}$\quad
Mubbasir Kapadia$^{2,4}$\quad
Yizhak Ben-Shabat$^{2}$\\
$^1$Australian National University\quad
$^2$Roblox\quad
$^3$Stanford University\quad
$^4$Rutgers University\\
{\tt\small $^1$\{jiaho.zhang, stephen.gould\}@anu.edu.au}\\
{\tt\small $^2$\{josephliu, ylee, smoon, vbzordan, haomiaojiang, ojacob, mkapadia, ibenshabat\}@roblox.com}\\
{\tt\small $^3$\{guytevet, karenliu\}@cs.stanford.edu}\\
{\tt\small\href{https://davidzhang73.github.io/romo-website}{https://davidzhang73.github.io/romo-website}}
}

\makeatletter
\let\oldmaketitle\@maketitle
\renewcommand{\@maketitle}{%
  \oldmaketitle%
  \begin{center}
    \vspace{-1em}
    \includegraphics[width=\linewidth]{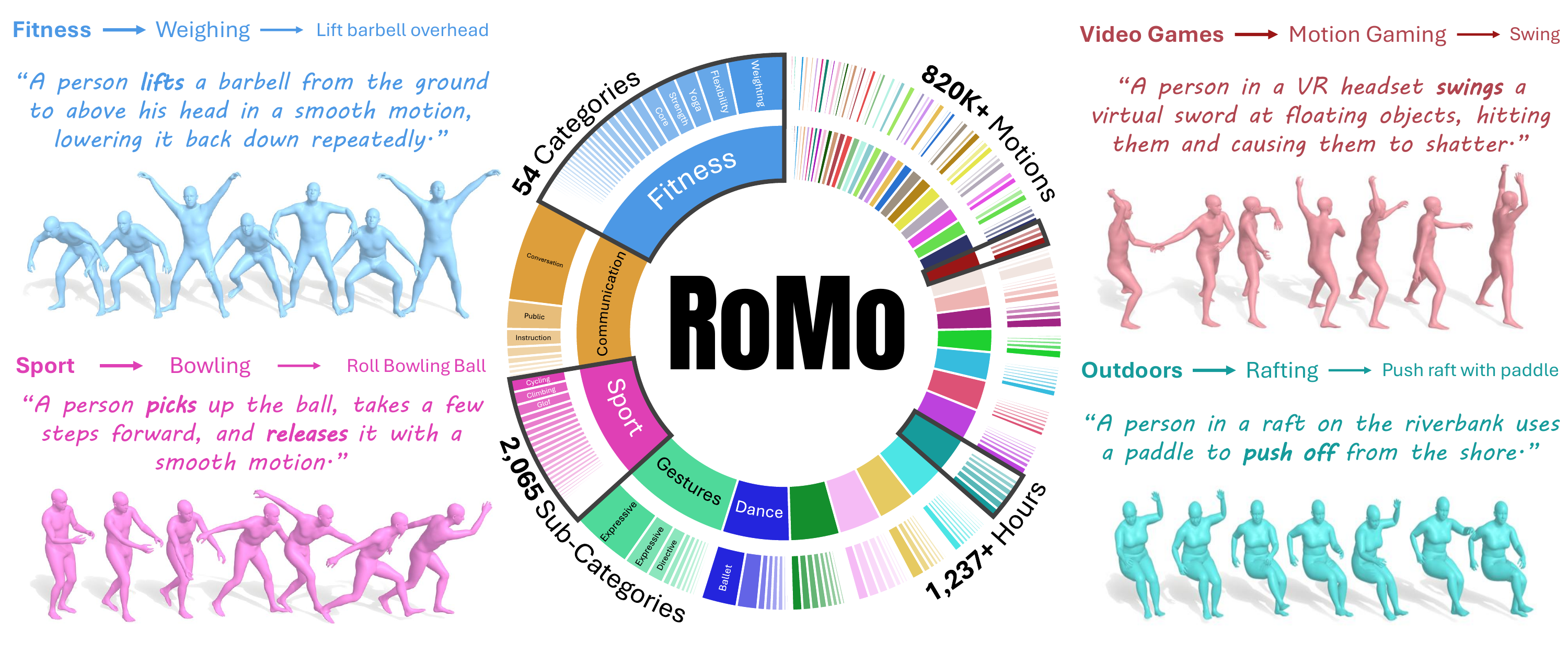}
    \captionof{figure}{
        We present \textbf{\OurDatasetName{}}, a large hierarchical dataset of 820K in-the-wild 3D human motions with detailed text captions organized into a three-level taxonomy (Category $\xrightarrow{}$ Subcategory $\xrightarrow{}$ Atomic-action), and annotated with text-rich prompts. The pie chart shows the distribution of categories and subcategories, while four examples illustrate diverse motions.
    }
    \label{fig:teaser}
  \end{center}
}
\makeatother

\begin{document}
\maketitle

%%%%%%%%%%%%%%%%%%%%%%%%%%%%%%%%
%%%%%%%%%%% Abstract %%%%%%%%%%%
%%%%%%%%%%%%%%%%%%%%%%%%%%%%%%%%

\begin{abstract}

Success in generative modeling across language, image, and video demonstrates that large, well-curated datasets are the key driver for building capable models. 3D Human motion, however, has lagged behind, constrained by an unsatisfying choice between small, high-fidelity motion capture datasets and large-scale in-the-wild collections dominated by static or low-quality sequences.

We introduce \textbf{\OurDatasetName{}}, a rich, large-scale, carefully curated dataset of in-the-wild human motions that resolves these tradeoffs. 
To ensure quality, we introduce a taxonomy-aware filtering pipeline that aggressively removes static and artifact-prone sequences. 
Every sequence is annotated with detailed captions and organized by a novel three-level semantic taxonomy. 
This hierarchical structure enables fine-grained, per-category evaluation, that reveals model strengths and weaknesses obscured by global metrics. 

We demonstrate that models trained on \OurDatasetName{} achieve state-of-the-art fidelity and diversity while gaining a superior understanding of complex, subtle text prompts. Finally, we release the Motion Toolbox to standardize metrics, data conversion, and visualization, establishing a foundation for reproducible and interpretable motion generation research.

\end{abstract}

%%%%%%%%%%%%%%%%%%%%%%%%%%%%%%%%%%%%
%%%%%%%%%%% Introduction %%%%%%%%%%%
%%%%%%%%%%%%%%%%%%%%%%%%%%%%%%%%%%%%

\section{Introduction}
\label{sec:intro}

\begin{table*}[htbp]
\centering
\caption{Comparison of \OurDatasetName{} with existing publicly available 3D motion datasets with free-form text annotations. \OurDatasetName{} is the first to integrate both a large-scale hierarchical semantic taxonomy and a massive-scale dataset, featuring 820K core clips (1237.8 hours).
For clarity, in the table: `Text diversity' refers to the number of captions per motion sequence. `Clip Number' reports both the core set (new motion sequences proposed in that work) and the total clips. 
}
\label{tab:dataset_comparison}
\resizebox{\textwidth}{!}{
\begin{tabular}{l c c rr c c ccc}
\toprule

\multirow{2}{*}{\textbf{Dataset}} & \multirow{2}{*}{\textbf{\begin{tabular}{@{}c@{}}Hierarchical \\ Semantic \\ Taxonomy\end{tabular}}} & \multirow{2}{*}{\textbf{\begin{tabular}{@{}c@{}} Category \\ Sub-category\end{tabular}}} & \multirow{2}{*}{\textbf{\begin{tabular}{@{}c@{}}Clip Number \\ Core (Total)\end{tabular}}} & \multirow{2}{*}{\textbf{Hour}} & \multirow{2}{*}{\textbf{Text Diversity}} & \multirow{2}{*}{\textbf{Source}} & \multicolumn{3}{c}{\textbf{Scene}} \\
\cmidrule(lr){8-10}

& & & & & & & \textbf{Indoor} & \textbf{Outdoor} & \textbf{RGB} \\[0.5em]
\midrule
KIT-ML~\cite{plappert2016kit}& \ding{55} & - &3.9K (3.9K) & 11.2 (11.2) & 1-3 & MoCap & \ding{51} & \ding{55} & \ding{55} \\
BABEL~\cite{punnakkal2021babel}& \ding{51} & 8 / 260 &13K (13K) & 43.5 (43.5) & 1 & MoCap & \ding{51} & \ding{55} & \ding{55} \\
HumanML3D~\cite{Guo_2022_CVPR}& \ding{55} & - &0 (15K) & 0 (28.6) & 1-3 & MoCap & \ding{51} & \ding{55} & \ding{55} \\
SnapMoGen~\cite{hwangsnapmogen}& \ding{55} & - &20K (20K) & 43.7 (43.7) & 6 & MoCap & \ding{51} & \ding{55} & \ding{55} \\
Motion-X~\cite{motionx}& \ding{55} & - &48.6K (81.1K) & 86 (144.2) & 1 & Video+MoCap & \ding{51} & \ding{55} & \ding{55} \\
Motion-X++~\cite{zhang2025motionx++}& \ding{55} & - &39.4K (120.5K) & 59 (180.9)& 1 & Video+MoCap & \ding{51} & \ding{55} & \ding{55} \\
% MotionLib~\cite{motionlib}& \ding{55} & 1.13M (1.21M) & 1456.4 & 2 & Video & \ding{51} & \ding{51} & \ding{51} \\
MotionMillion~\cite{motionmillion}& \ding{55} & - &560K (2M) & 726.5 (2000) & $>$20 & Video & \ding{51} & \ding{51} & \ding{51} \\
\midrule
\textbf{\OurDatasetName{} (ours)} & \textbf{\ding{51}} & \textbf{54 / 2065} & \textbf{820K (2.58M)} & \textbf{1237.8 (3023)} & 5 & Video & \ding{51} & \ding{51} & \ding{51} \\
\bottomrule
\end{tabular}
} % resizebox
\end{table*} 

3D human motion generation has advanced rapidly with the rise of diffusion~\cite{ho2020denoising,tevet2023human} and GPT~\cite{radford2019language,guo2022generating} models, enabling high-fidelity synthesis and a wide range of controllable behaviors. At the foundation of this progress lies the dataset. Learning from the scaling success in language~\cite{grattafiori2024llama}, image~\cite{flux2024}, and video~\cite{wan2025} generation, the field has attempted to move beyond small, high-fidelity motion capture collections by extracting poses from in-the-wild videos. In practice, however, these pioneering large-scale efforts suffer from minimal curation, resulting in datasets dominated by static, noisy, or artifact-prone sequences. This has led to an unsatisfying choice: train on small, clean datasets that no longer challenge modern models, or on massive but unreliable collections that bias models toward static, low-quality motion. This tradeoff limits progress, as neither option provides the reliable, fine-grained data needed to train or evaluate truly compositional human motion.

To address these challenges, we introduce \textbf{\OurDatasetName{}}, a large-scale collection of in-the-wild 3D human motion paired with rich textual and categorical annotations. The dataset is built on three pillars: \emph{scale}, \emph{curation}, and a \emph{coarse-to-fine taxonomy}. Together, these components create a foundation for reliable training and transparent evaluation in motion generation.

First, we gather an unprecedented volume of videos depicting diverse human activities from multiple online sources. Each clip is processed with a SOTA pose estimation model, GVHMR~\cite{shen2024gvhmr}, to extract accurate 3D motion sequences. To enrich the semantic context, we employ vision-language models to generate multiple detailed captions per clip, capturing both the action and its environment.

Second, we place strong emphasis on curation. Instead of releasing raw, noisy data, we apply a multi-stage filtering pipeline guided by quantitative motion metrics to remove static, incomplete, or artifact-prone samples. The filtering process is adaptive: for example, motions in the ``fishing'' category are expected to be more subtle than those in ``acrobatics'' and the threshold is adjusted accordingly to retain only realistic, category-consistent sequences.

Finally, we propose a new hierarchical taxonomy for human motion, organizing each sequence into categories, subcategories, and atomic actions. This taxonomy introduces a structured way to analyze and evaluate motion generation models. It enables researchers to assess performance per category, revealing specific strengths, weaknesses, and blind spots, while also ensuring balanced coverage of the human motion space and mitigating dataset bias.

To further advance reproducibility and accessibility, we release 
the \textbf{Motion Toolbox}, a unified framework for data conversion, standardized evaluation metrics, and browser-based visualization. Together, these contributions establish a new foundation for large-scale, transparent, and semantically grounded research in generative human motion. 

We demonstrate that RoMo drives substantial gains in fidelity, diversity, and prompt understanding. 
Contemporary motion generation models~\cite{tevet2023human,motionmillion} trained on our benchmark respond to subtle textual nuances and generalize to scenarios out of distribution in prior datasets. Leveraging our taxonomy, we perform the first per-category evaluation of generative models, revealing that state-of-the-art models, while excelling at common actions, fail on fine-grained categories with subtle interactions.
This union of data scale, motion quality, rich text prompts, and taxonomy structured evaluation surfaces research gaps that previous datasets could not reveal and lays the groundwork for the next stage of human motion generation.

\noindent The main contributions of this paper are:
\begin{itemize}
    \item \textbf{\OurDatasetName{}:} A new large-scale, in-the-wild motion dataset (1237h, 820K clips) with five rich text captions per clip and high-quality, filtered motion.
    \item \textbf{A Novel Curation \& Taxonomy Framework:} We introduce a hierarchical semantic taxonomy (54 categories, 2,065 subcategories) and a \textit{taxonomy-aware adaptive filtering pipeline} that uses it to ensure high motion quality and diversity.
   \item \textbf{The Motion Toolbox:} An open-source library for standardized evaluation, data conversion, and browser-based visualization to accelerate reproducible research.
\end{itemize}

%%%%%%%%%%%%%%%%%%%%%%%%%%%%%%%%%%%%
%%%%%%%%%%% Related Work %%%%%%%%%%%
%%%%%%%%%%%%%%%%%%%%%%%%%%%%%%%%%%%%

\section{Related Works}
\label{sec:related}

Table~\ref{tab:dataset_comparison} presents a comparison of publicly available human motion datasets, from motion capture and video sources.

\textbf{Human Motion Generation.}
In recent years, human motion generation has made remarkable progress, driven by the adaptation of modern machine learning paradigms. ACTOR~\cite{petrovich2021actor} and MotionCLIP~\cite{tevet2022motionclip} introduced a transformer-based auto-encoder for effective motion synthesis \cite{vaswani2017attention}. MDM~\cite{tevet2023human} advanced the field by applying diffusion models~\cite{ho2020denoising}, enabling diverse and high-fidelity text-to-motion generation. Building on this direction, DiP~\cite{tevet2025closd} and CAMDM~\cite{camdm} brought diffusion to real-time performance.
In parallel, inspired by the success of GPT~\cite{radford2019language} models in language generation, several works~\cite{guo2022generating,jiang2024motiongpt,zhang2023generating} demonstrated that human motion can be tokenized and generated as a sequence of discrete motion tokens~\cite{van2017neural}. MoMask~\cite{guo2024momask} introduced a residual quantization to enhance fidelity.

These generative approaches have been successfully applied to a wide range of animation tasks, including text-to-motion~\cite{jiang2024motiongpt,meng2024rethinking}, music-driven motion~\cite{alexanderson2023listen,siyao2022bailando,lee2019dancing,tseng2023edge,dabral2023mofusion}, motion stylization~\cite{sawdayee2025dance,zhong2024smoodi}, multi-person~\cite{shafir2024human,liang2024intergen}, and human–object interaction~\cite{pi2025coda,li2023object,peng2023hoi,ron2025hoidini,li2024task,zhang2025bimart}. Alongside these advances, a variety of techniques for fine-grained control have emerged, leveraging diffusion guidance~\cite{karunratanakul2023gmd}, inpainting~\cite{shafir2024human,cohan2024flexible}, noise optimization~\cite{karunratanakul2024optimizing}, and attention injections~\cite{raab2024monkey}.
However, this entire line of research mostly relies on a small set of motion capture datasets. This constrains these powerful models, limiting their ability to learn the diversity and nuance of complex, in-the-wild human motion—a gap we directly address with \OurDatasetName{}.

\textbf{Motion Capture Collections.}
Motion capture (mocap) has long served as the gold standard for recording 3D human motion in character animation. Using wearable sensors~\cite{DIP:SIGGRAPHAsia:2018}, optical markers~\cite{ionescu2013human3,mandery2015kit}, or calibrated multi-view systems~\cite{joo2015panoptic}, mocap provides highly accurate reconstructions of body movement. AMASS~\cite{Mahmood2019AMASSAO} unified a collection of academic mocap datasets \cite{
AMASS_ACCAD,
AMASS_BMLhandball,
AMASS_BMLmovi,
AMASS_BMLrub,
AMASS_CMU,
AMASS_DanceDB,
AMASS_DFaust,
AMASS_EyesJapanDataset,
AMASS_GRAB,
AMASS_GRAB-2,
AMASS_HDM05,
AMASS_HUMAN4D,
AMASS_HumanEva,
AMASS_KIT-CNRS-EKUT-WEIZMANN,
AMASS_KIT-CNRS-EKUT-WEIZMANN-2,
AMASS_KIT-CNRS-EKUT-WEIZMANN-3,
AMASS_MOYO,
AMASS_MoSh,
AMASS_PosePrior,
AMASS_SFU,
AMASS_SOMA,
AMASS_TCDHands,
AMASS_TotalCapture,
AMASS_WheelPoser} 
by converting them into a consistent SMPL~\cite{SMPL:2015} representation, establishing a common format for research. BABEL~\cite{punnakkal2021babel} extended this effort with basic textual annotations, while HumanML3D~\cite{Guo_2022_CVPR} curated a refined subset of AMASS with detailed textual descriptions, making it the most widely used dataset for text-to-motion generation. Additional collections such as AIST++~\cite{tsuchida2019aist} captured around five hours of dance motion paired with music, while GRAB~\cite{taheri2020grab}, OMOMO~\cite{li2023object}, and HUMOTO~\cite{Lu_2025_HUMOTO} introduced object interactions, hand articulation, and multi-person motions, each annotated with textual action descriptions.
However, mocap remains an expensive and time-consuming process, and even the combined output of academic collaborations around the globe amounts to fewer than 20K motions~\cite{Mahmood2019AMASSAO}.
This small scale, in stark contrast to the 400M images in LAION~\cite{schuhmann2021laion}, makes it impossible for MoCap-based datasets to capture the `long tail' of diverse human activities that \OurDatasetName{} is designed to represent.

\textbf{Motions from Videos In-the-wild.}
Recent progress in 3D human and camera pose estimation~\cite{shen2024gvhmr,shin2024wham,ye2023decoupling} has opened the door to extracting motion directly from monocular in-the-wild videos. This raises the question of whether it can be leveraged to break the scale limitations of motion capture. 
Motion-X~\cite{motionx, zhang2025motionx++} and MotionMillion~\cite{motionmillion} took steps in this direction, releasing large datasets with up to a million motion sequences. Motionlib \cite{motionlib} also curated a large-scale dataset, but as it has not been publicly released, it is omitted from our comparative analysis. Despite the scale of these collections, data quality remains a significant issue, as pose estimation still suffers from significant errors: inaccurate camera reconstruction can cause characters to float, temporal inconsistencies lead to jitter, and regression noise often produces anatomically implausible poses.
Moreover, many collected clips are static or lack meaningful motion. As a result, these large-scale datasets, while impressive in size, often include numerous artifacts and imbalanced motion categories, underscoring the need for stronger filtering and structured organization mechanisms.
This underscores the critical need for stronger filtering and structured organization. \OurDatasetName{} is the first dataset designed to solve both problems: our taxonomy-aware adaptive filtering pipeline addresses the quality issue, while our hierarchical taxonomy solves the lack of structured organization.

\subsubsection{Human Motion Evaluation.}
Many works have proposed objective measures for motion quality, such as foot skating~\cite{ron2025hoidini, tevet2025closd, yuan2023physdiff, karunratanakul2023gmd}, floating \cite{tevet2025closd, yuan2023physdiff}, jitter (jerk)~\cite{shen2024gvhmr, mu2025stablemotion, motionmillion, barquero2024flowmdm}, ground~\cite{ron2025hoidini, yuan2023physdiff} or self-penetration~\cite{muller2021self}, and root orientation change \cite{motionmillion},  yet each study defines its metrics slightly differently, leaving no clear consensus. HumanML3D~\cite{Guo_2022_CVPR} established a standard for evaluating generated motion distributions by introducing neural evaluators that embed text and motion in a shared latent space to compute FID~\cite{heusel2017gans} and prompt adherence. SnapMoGen~\cite{hwangsnapmogen} later refined these evaluators.
Our Motion Toolbox (\secref{sec:toolbox}) is designed to provide a single, open-source, and verified standard implementation of these key metrics.

%%%%%%%%%%%%%%%%%%%%%%%%%%%%%%%%
%%%%%%%%%%% Pipeline %%%%%%%%%%%
%%%%%%%%%%%%%%%%%%%%%%%%%%%%%%%%

\begin{figure*}[t]
  \centering
  \includegraphics[width=\linewidth]{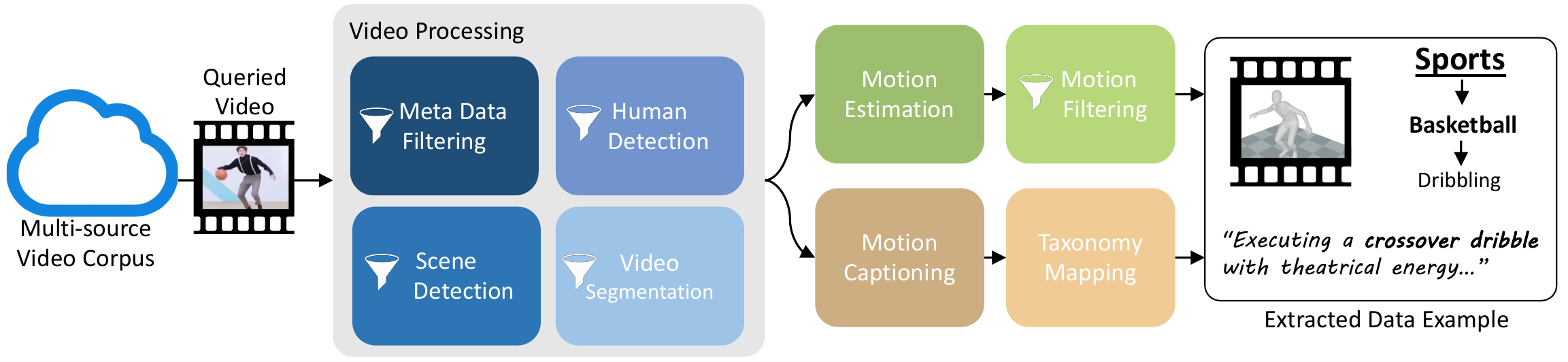}
  \caption{
  \textbf{Data Pipeline.} \OurDatasetName{} is extracted from a large web video corpus.
    We query human motion videos, filter single-human scenes, and segment them into atomic actions. We then apply a 3D camera and pose estimation, remove low-quality motions, and caption and categorize the results using our hierarchical taxonomy.
    Overall, the pipeline performs uncompromising large-scale filtering that processes 125K hours of raw footage and distills 1\% into high-quality, well-annotated motions.
  }
  \label{fig:pipeline}
\end{figure*}

\section{Taxonomy‑aware Data Pipeline}
\label{sec:method}

We present our fully automated, taxonomy-aware data pipeline in \figref{fig:pipeline}. 
This system extracts diverse human motions from broad web video corpora while preserving semantic coverage and motion quality.

\OurDatasetName{} stands on two strong foundations: a hierarchical taxonomy of the human motion action space (\secref{sec:tax}) and a multi-source video corpus (\secref{sec:data_query}).
The taxonomy is a hierarchical protocol that spans the human motion manifold and organizes it into categories, subcategories, and atomic actions, providing the structural backbone of the dataset.
The video corpus combines large-scale web queries with public datasets, yielding a diverse and massive pool of candidate clips that cover a wide spectrum of everyday and specialized motions.

The video processing module (\secref{sec:video}) filters the queried videos based on their metadata, splits them into scenes, identifies scenes containing a single moving person, and then segments each scene into atomic action candidates.
The filtered clips undergo 3D body and camera motion estimation (\secref{sec:estimate}), followed by evaluation and filtering metrics that remove static or corrupted reconstructions (\secref{sec:motion_eval}).
In parallel, a vision-language model then generates descriptions of the human motion, using the visual clues in the surrounding video context to refine and disambiguate the motion phrasing, and a taxonomy-mapping stage assigns each segment to its category, subcategory, and atomic action.

\begin{figure}
    \centering
    \includegraphics[width=0.98\linewidth]{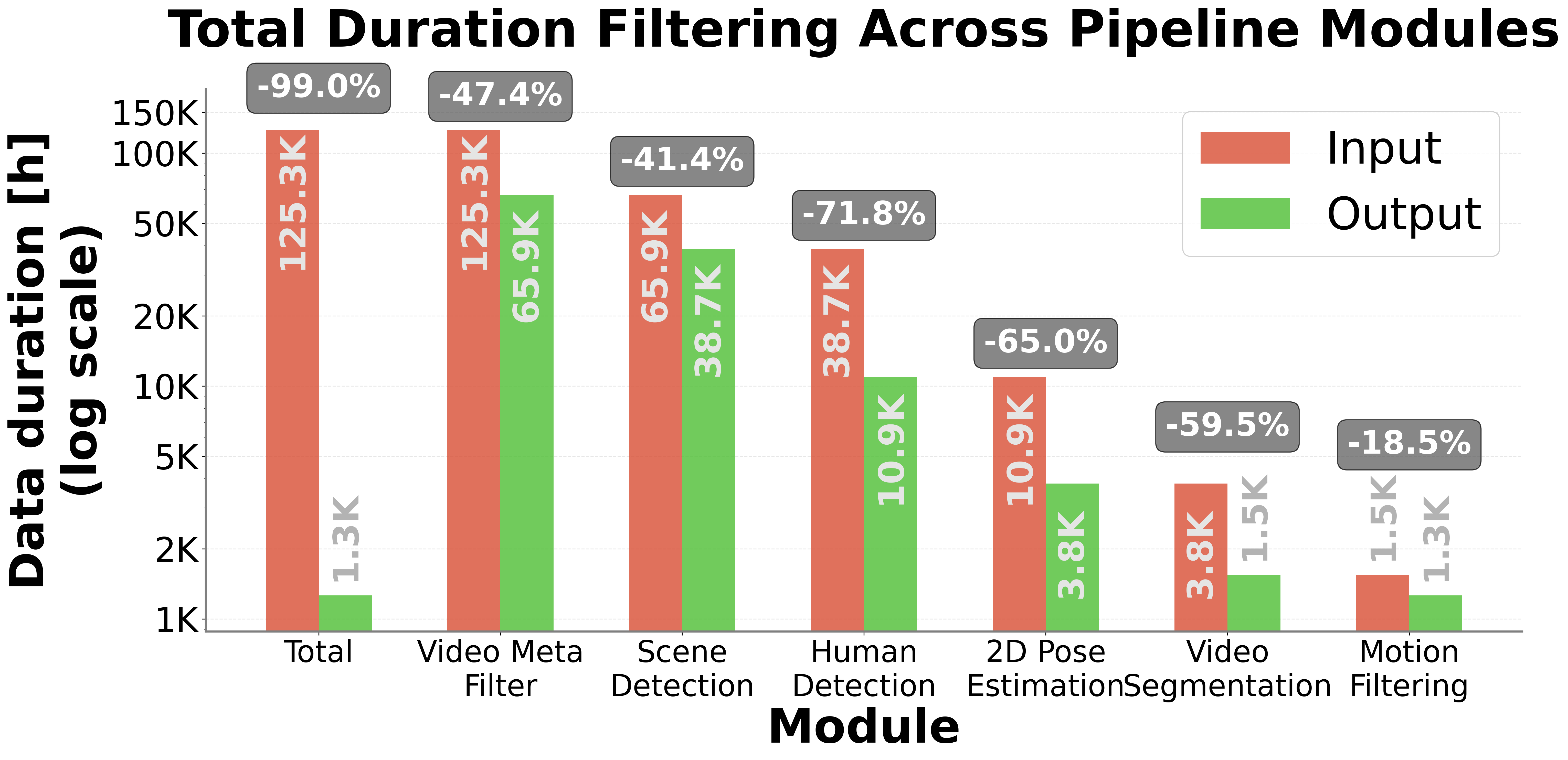}
    \caption{\textbf{Aggressive filtering.}  Filtering out 99\% of total input duration in our data pipeline. The chart shows the input (red) and output (green) hours for each filtering module, demonstrating a reduction from 125.3K to 1.3K total hours.}
    \label{fig:total_data_reduction}
\end{figure}

Overall, the system applies aggressive filtering at a massive scale (\figref{fig:total_data_reduction}), processing 125K hours ($\sim$14 years) of human motion videos and distilling only about one percent into 3D motion sequences of uncompromising quality with taxonomy-aligned annotations.

\subsection{Motion Taxonomy}
\label{sec:tax}

We introduce a three-level motion taxonomy to ensure broad coverage of human activities and to support evaluation at both coarse and fine semantic resolutions. The initial structure draws on action recognition literature \cite{tang2019coin, ben2021ikea, caba2015activitynet} and was expanded to cover the diversity observed in large web corpora. This design enables hierarchical analysis. For example, if a generative model struggles with the \textit{gestures} category, the taxonomy allows drilling down to specific gesture types to identify failure modes.

The taxonomy contains three layers (Category $\xrightarrow{}$ Subcategory $\xrightarrow{}$ Atomic-action). 
\textbf{Categories} capture broad motion themes such as \emph{Sports}, \emph{Daily Activities}, and \emph{Professions}. 
\textbf{Subcategories} form compact semantic groups expressed as noun phrases, for example \emph{Table Tennis} or \emph{Cleaning Activities}. 
\textbf{Atomic-actions} are short present-tense verb phrases that may include objects or body parts but avoid modifiers and punctuation, for instance \emph{Swing racket} or \emph{Climb stairs}. These units span only a few seconds and serve as the basis for segmentation, search, and captioning.

We construct the taxonomy with a hybrid approach. High-level categories were defined through a top-down review of major action and video datasets \cite{tang2019coin, ben2021ikea, caba2015activitynet, liu2025hoigen, wang2025videoufo, li2025openhumanvid, nan2024openvid}. Then, 2,897 subcategories and 28,874 atomic actions were expanded through LLM-assisted term discovery, followed by human curation to refine phrasing and merge duplicates. 
The final structure contains 54 categories, 2,065 subcategories, and 28,874 of atomic actions. A full listing appears in the supplemental material.

\subsection{Taxonomy-based Video Queries}
\label{sec:data_query}

The taxonomy structure serves as the basis for querying the video corpus looking for human motion videos:
given a Category–Subcategory pair together with its Atomic Action vocabulary, an LLM synthesizes $N$ diverse search queries that cover synonyms, salient objects, and contexts; each query retrieves up to $M$ candidate videos from online platforms, yielding as many as $N\!\times\!M$ candidates per Subcategory after de‑duplication via URL hashing and perceptual video fingerprints.
We further incorporate publicly available video datasets \cite{carreira2017kinetics, wang2025videoufo, li2025openhumanvid, liu2025hoigen} and unify them to our taxonomy post‑hoc: free‑form labels or titles are first converted to normalized textual descriptions and then mapped to atomic-actions with a retrieval‑augmented LLM remapper constrained by MCP (synonym tables, type checks, and rationale logging). 

\subsection{Video Processing and Filtering}
\label{sec:video}

\textbf{Metadata Filtering.}
We first observe that a large fraction of queried videos can be discarded using metadata alone. We first remove clips that are too short or have a frame rate below 24 FPS. 
An LLM then reads the remaining metadata (mainly tags and descriptions) and assigns confidence scores in the range $[0,1]$
to four criteria: (1) the description refers to a human action, (2) the action involves a single person, (3) the full body is expected to be visible, and (4) the content is not AI generated. Videos that fall below the threshold on any criterion are filtered out.

\textbf{Scene Detection.}
To ensure each candidate clip contains exactly one visible human performing an analyzable motion, we follow MotionMillion~\cite{motionmillion} by first applying \texttt{PySceneDetect} to remove transitions and discontinuous segments. In addition, our method further eliminates near‑static sequences based on inter‑frame differences. 

\textbf{Single-human Detection.}
We then detect humans with YOLOv8~\cite{yolov8_ultralytics} and retain clips in which a single human is present in a large fraction of frames and occupies a reasonable portion of the image; extreme close‑ups and tiny subjects are discarded. 
Finally, we estimate 2D poses with ViTPose~\cite{xu2022vitpose} and reject clips with frequent truncation or low in‑frame joint ratios, thereby improving downstream motion recovery robustness. 
The resulting clips range from seconds to minutes and are visually continuous, each dominated by a single performer.

\textbf{Temporal Semantic Segmentation.}
A key limitation of prior large-scale motion datasets is their use of fixed-length slicing or hard duration caps, which often breaks clips at points that do not correspond to meaningful actions. 
MotionMillion~\cite{motionmillion} for example, inherit this issue since their segmentation is not aligned to semantic units. 
To address this, we perform \emph{semantic} temporal segmentation using a state-of-the-art multimodal VLM (Qwen3-VL~\cite{qwen3}). 
For each cleaned clip, the model receives the relevant taxonomy node together with the permissible atomic-action vocabulary and returns a sequence of segments, each with an action label and precise start and end times. We gate all labels through the Subcategory vocabulary with synonym resolution, merge gaps and very short fragments to stabilize boundaries, and clamp timestamps to the clip duration. This produces atomic action spans aligned with human intent, surpassing the quality of uniform or fixed-length slicing.

\subsection{Motion Estimation and Descriptions}
\label{sec:estimate}

Upon identifying an atomic-action segment, the clip is processed through two concurrent modules.

The first module estimates 3D human motion using GVHMR \cite{shen2024gvhmr}, producing outputs in the SMPL \cite{SMPL:2015} body model format, which include global translation, root orientation, and 24 joint angles. 
These motion sequences are then resampled to 30FPS, standardized in body scale, and orientation-aligned to ensure cross-instance comparability.

Simultaneously, the second module extracts text descriptions by utilizing Qwen3-VL \cite{qwen3} to generate $K$ natural language descriptions for the motion sequence.
Following this, these outputs undergo an additional processing stage, again using Qwen3-VL, to extract structured taxonomy labels (category, subcategory, and atomic action).
This final step ensures that the extracted labels align with the provided taxonomy for inputs that have one, and generates new, consistent labels for those that do not (from video datasets).

\subsection{Motion Evaluation and Filtering}
\label{sec:motion_eval}

\begin{figure}
    \centering
    \includegraphics[width=0.9\linewidth]{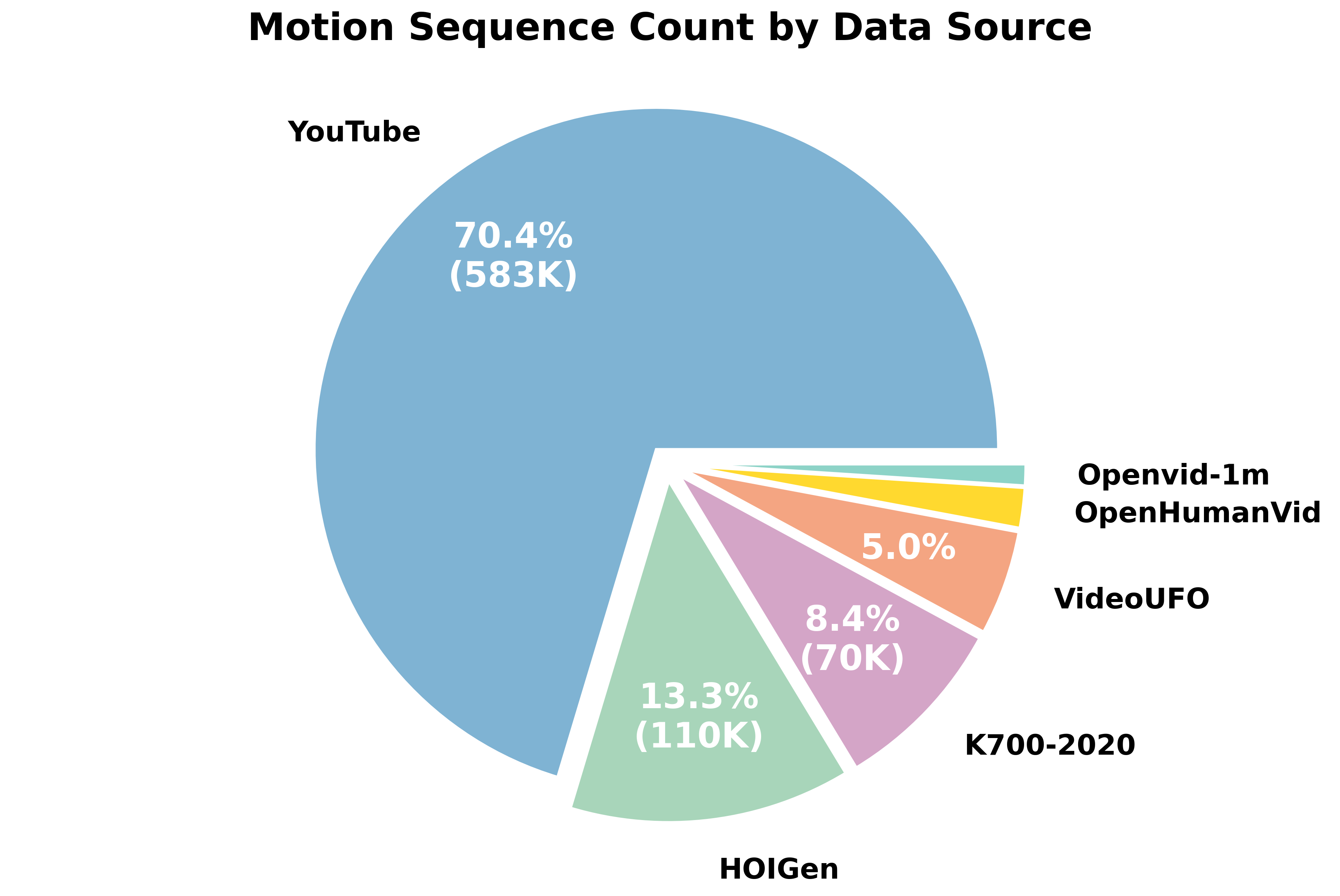}
    \caption{\textbf{Data source distribution for \OurDatasetName{}}. The pie chart illustrates the percentage breakdown of motion sequences sourced from web videos}
    \label{fig:data_source_distribution}
\end{figure}

When extracting motion from web videos, analyzing the motion's level of dynamics has been grossly overlooked, a gap that, as we show, has resulted in datasets dominated by static or low-activity motions.
To address this, we propose the \emph{Dynamic Score}. 
The dynamic score quantifies motion activity by combining temporal and spatial characteristics. 

Given a motion sequence with joint positions $\mathbf{P} \in \mathbb{R}^{F \times J \times 3}$ and joint velocities $\mathbf{V} \in \mathbb{R}^{F \times J \times 3}$ where $F$ is the number of frames and $J$ is the number of joints, we compute two components. The \textit{temporal component} measures instantaneous motion activity through velocity magnitudes:
\begin{equation}
S_{\text{temporal}} = \frac{1}{F \cdot J} \sum_{t=1}^{F} \sum_{j=1}^{J} \|\mathbf{v}_{t,j}\|_2
\end{equation}
The \textit{spatial component} captures the overall extent of motion by measuring the range of each joint's trajectory:
\begin{equation}
S_{\text{spatial}} = \frac{1}{J} \sum_{j=1}^{J} \left\| \max_{t} \mathbf{p}_{t,j} - \min_{t} \mathbf{p}_{t,j} \right\|_2
\end{equation}
These components are combined using weights $w_v$ and $w_r$ to produce the final dynamic score:
\begin{equation}
S_{\text{Dynamic}} = w_v \cdot S_{\text{temporal}} + w_r \cdot S_{\text{spatial}}
\end{equation}
This hybrid approach ensures that both highly dynamic motions (e.g., dance, sports) and motions with large spatial coverage (e.g., reaching, walking) receive appropriate scores. Throughout the paper we use $(w_v, w_r) = (0.7, 0.3)$.

The dynamic score serves as our last-stage filter, allowing us to exclude motions that lack meaningful activity. 
Since different types of actions have inherently different dynamism levels (for example, gymnastics compared to screwing in a light bulb), we suggest an \emph{adaptive per-category filtering} rather than a universal threshold. Selecting the top-P percentile within each category guarantees that even subtle activities remain well represented, while overly static clips are removed.

%%%%%%%%%%%%%%%%%%%%%%%%%%%%%%%%%%
%%%%%%%%%%% Dataset %%%%%%%%%%%
%%%%%%%%%%%%%%%%%%%%%%%%%%%%%%%%%%

\section{The \OurDatasetName{} Dataset}
\label{sec:our_dataset}

\OurDatasetName{} is an extensive 3D human motion dataset, sourced from a diverse collection of web videos and existing open-source datasets.
It comprises 813,938 motion sequences that total $\sim$1,238 hours of human motion at 30FPS. 
The clips have an average length of 165 frames (median 114), with all sequences capped to a duration between 30 and 600 frames. Each sequence is accompanied by five text descriptions and our taxonomy category, subcategory, and atomic action labels (see \figref{fig:data_source_distribution} for the data source distribution). 
As demonstrated in \tabref{tab:dataset_comparison}, \OurDatasetName{} is uniquely characterized by its hierarchical semantic taxonomy—a feature absent in existing datasets. 
While achieving a scale competitive with modern video-based collections, it far surpasses the diversity of traditional mocap data. 
Note that to ensure a fair comparison, we distinguish between the 'total set' (all reported sequences) and the 'core set' (newly introduced sequences). 
This separation is crucial because common practice often involves merging previous datasets, which obscures the actual number of novel motions contributed by a specific paper.

\begin{figure}[tb]
    \centering
    \begin{subfigure}[b]{\linewidth}
        \centering
        \includegraphics[width=0.8\linewidth]{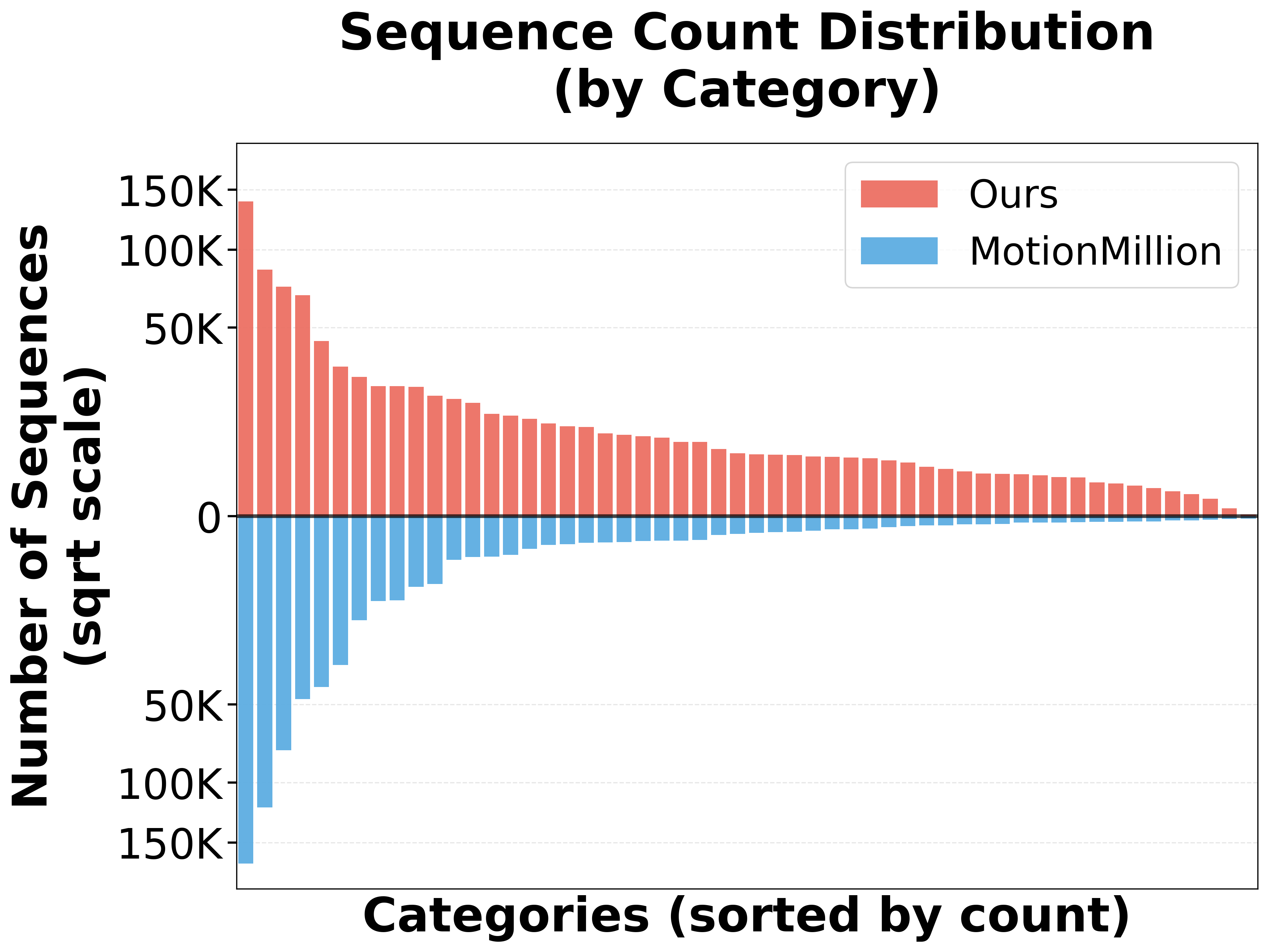}
        \caption{}
        \label{fig:sequence_count_bidirectional}
    \end{subfigure}
    
    \vspace{0.15em}
    
    \begin{subfigure}[b]{0.8\linewidth}
        \centering
        \includegraphics[width=\linewidth]{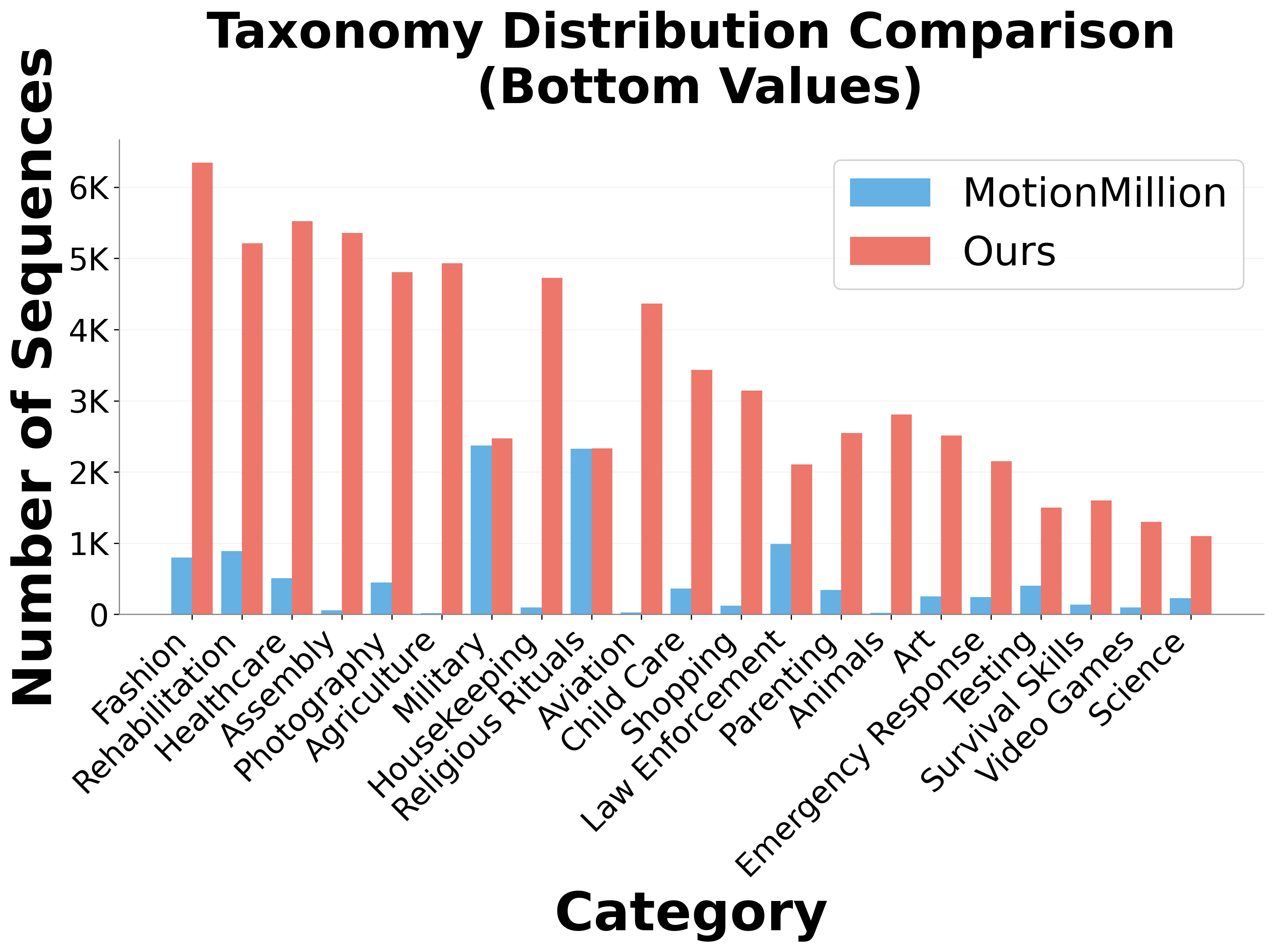}
        \caption{}
        \label{fig:sequence_count_tail}
    \end{subfigure}
    \caption{\textbf{Superior diversity and coverage}. Comparison of sequence counts per category between our \OurDatasetName{}  and MotionMillion (a). The bottom figure (b) shows the ``tail'' of the distribution of both datasets on the same plot, demonstrating how our dataset provides better coverage of these less frequent categories.}
    \label{fig:sequence_count}
\end{figure}

By organizing our data using this proposed taxonomy, we identify 54 categories and 2,065 distinct subcategories. 
When mapped to the same taxonomy, MotionMillion covers only 1,277 subcategories, marking a 61.7\% increase in coverage for our dataset. 
\figref{fig:sequence_count} illustrates this superior diversity by comparing our sequence distribution against MotionMillion's core set across all categories (a) and least common categories (b). 
For a comprehensive analysis at the subcategory level, please refer to the interactive HTML in the supplemental material.

\textbf{Dynamic Score.} 
A key contribution of our work is recognizing that motion datasets have long overlooked the distribution of motion dynamics, leading to inflated counts dominated by low-activity clips.
Our analysis reveals a significant portion of moderate-to-low dynamic motions in existing data. For instance, when filtering the 559K sequences from the MotionMillion dataset by dynamic score thresholds $\ge$0.05, $\ge$0.10, $\ge$0.15, and $\ge$0.50, only 88.41\%, 78.46\%, 69.07\%, and 31.55\% of sequences are retained, respectively. This highlights a heavy skew towards low-dynamic content.

\begin{figure}
    \centering
    \includegraphics[width=0.85\linewidth]{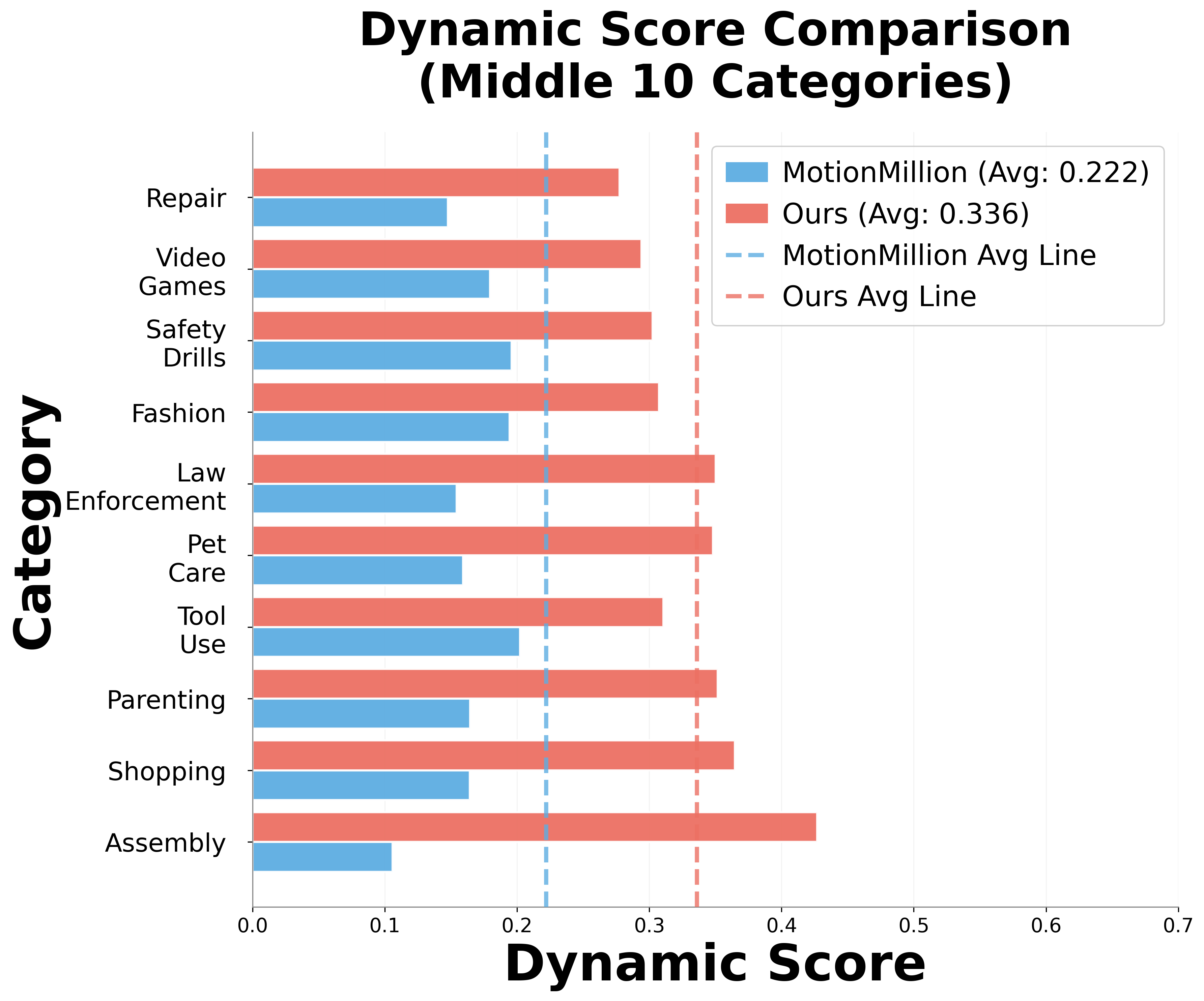}
    \caption{\textbf{Dynamic Score Analysis.} \OurDatasetName{} demonstrates higher dynamic scores across the majority of categories, with a 41\% improvement in dynamic score. This figure shows a subset of 10 categories. (For all categories, see supplemental material.) }
    \label{fig:comparison_dynamic_score}
\end{figure}

In contrast, our dataset is demonstrably more dynamic, achieving a mean dynamic score of 0.336, which is 41.4\% higher than MotionMillion's 0.222. This per-category advantage is further detailed in \figref{fig:comparison_dynamic_score}, which illustrates that our dataset provides more dynamic motions across multiple categories.
An extended per-category breakdown is provided in the supplemental material and reveals that categories inherently possess a broad range of dynamic scores (\eg, dance vs. eating), suggesting that models trained on such data must learn these differences implicitly.

\begin{figure}[t]
    \centering
    \begin{minipage}[c]{0.495\linewidth}
        \centering
        \includegraphics[width=\linewidth]{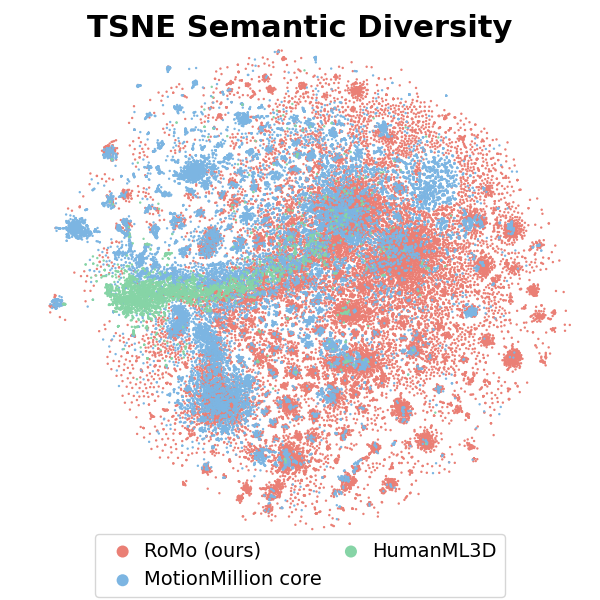}
    \end{minipage}
    \hfill 
    \begin{minipage}[c]{0.495\linewidth}
        \centering
        \includegraphics[width=\linewidth]{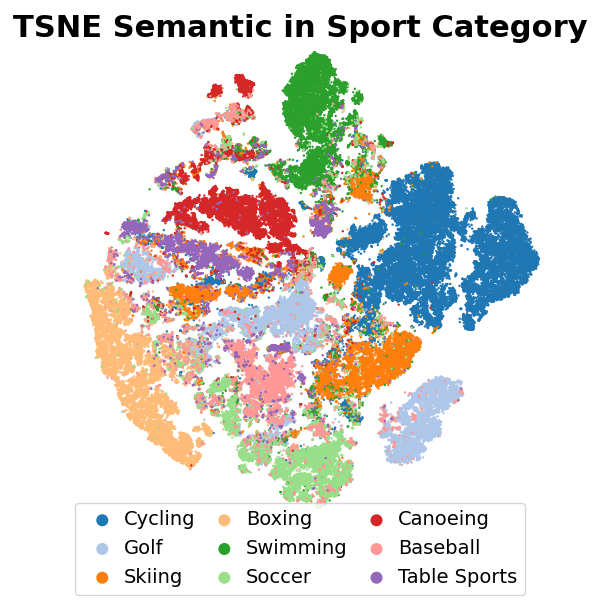}
    \end{minipage}
    \caption{\textbf{t-SNE analysis.} (left) Comparison of \OurDatasetName{} (ours) and MotionMillion and HumanML3D, showing improved coverage. (right) Semantic clustering of 'Sports' category, where points are colored by subcategory, confirming our Taxonomy's quality.}
    \label{fig:tsne}
\end{figure}

\textbf{Coverage and diversity.}  
To assess the semantic coverage and diversity of \OurDatasetName{}, we performed a t-SNE analysis on its text captions (see \figref{fig:tsne}). 
A qualitative comparison against MotionMillion and HumanML3D demonstrates that our dataset achieves better coverage and better diversity. 
Furthermore, we show internal semantic structure using our 'Sports' category, coloring each point by its subcategory (\eg, 'swimming,' 'golf,' 'baseball'). 
The results reveal distinct, semantically meaningful clusters, which confirm that our dataset's taxonomy successfully captures the hierarchical relationships between motions.

\section{The Motion Toolbox} 
\label{sec:toolbox}
\begin{figure}[tb]
    \centering
    \begin{minipage}{0.85\columnwidth}
    % --- Top Row (Swapped) ---
    % [t] aligns the subfigures at their top
    \begin{subfigure}[t]{0.48\columnwidth}  % <-- CHANGED [b] to [t]
        \centering
        \includegraphics[width=\linewidth]{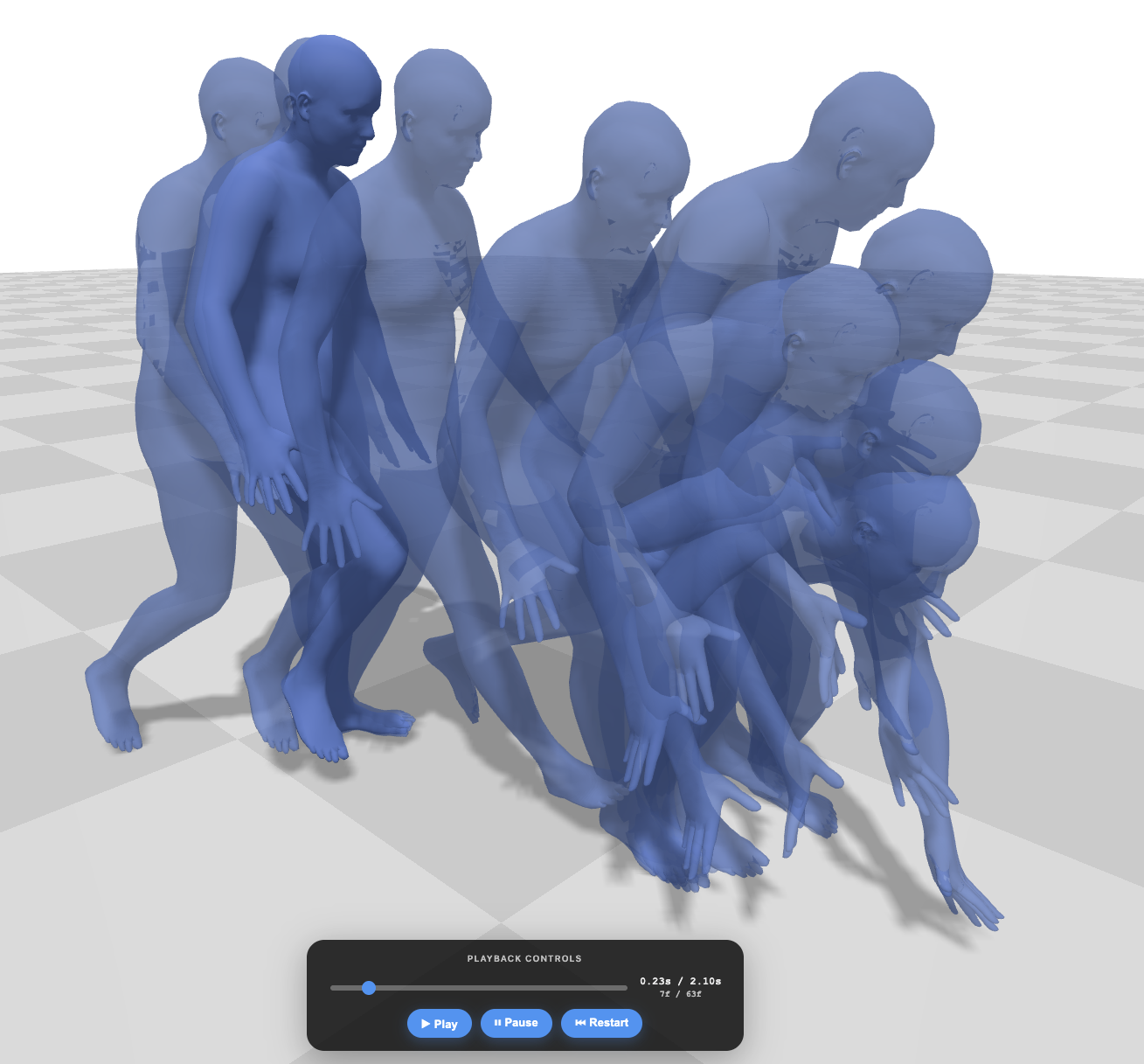}
        % \caption{} % Will be (a)
        \label{fig:top-left-new} % New label
    \end{subfigure}% <--- ADDED % to prevent extra horizontal space
    \hfill % Adds flexible horizontal space
    \begin{subfigure}[t]{0.48\columnwidth}  % <-- CHANGED [b] to [t]
        \centering
        \includegraphics[width=\linewidth]{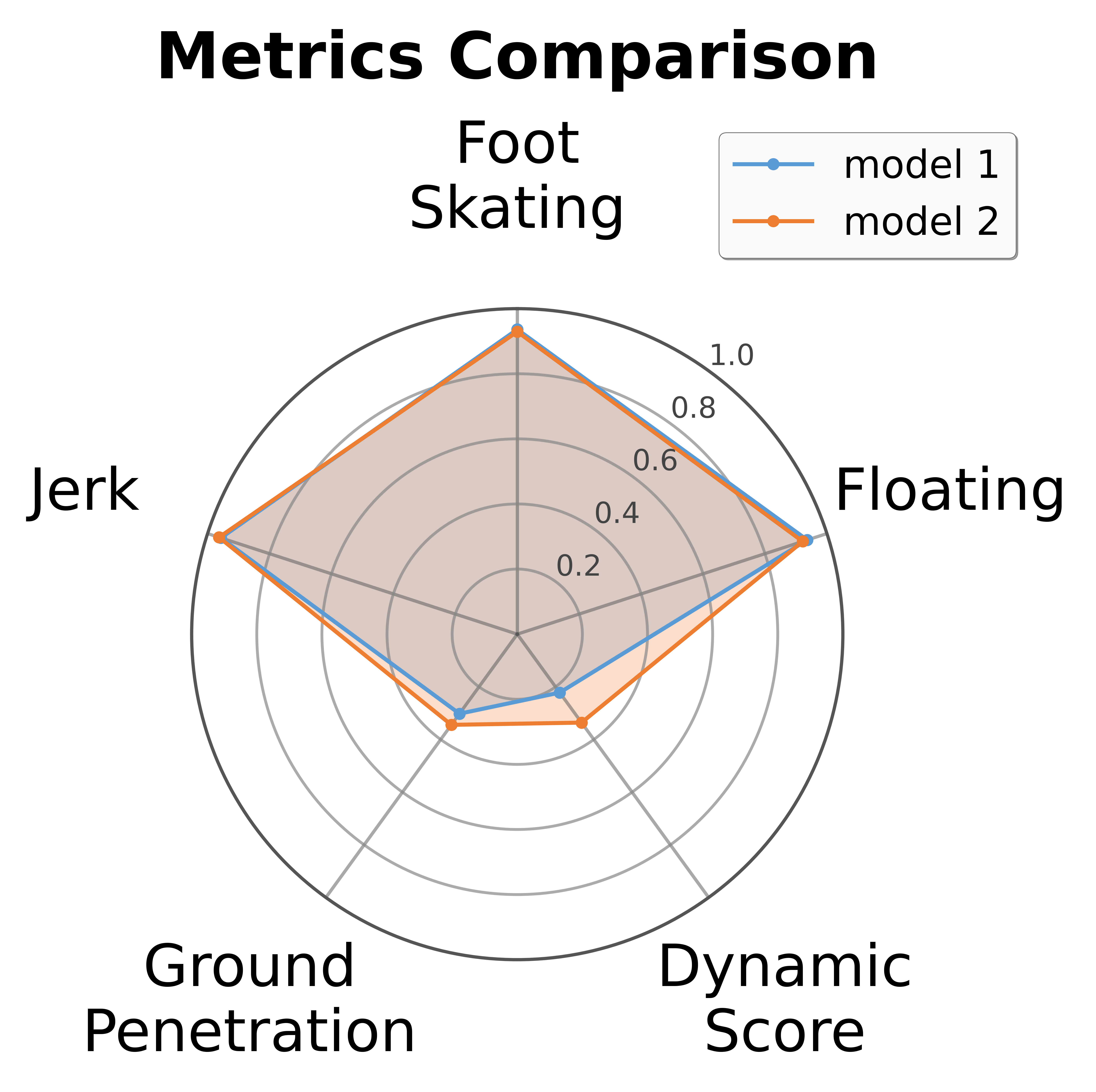}
        % \caption{} % Will be (b)
        \label{fig:top-right} 
    \end{subfigure}

    % Reduced vertical space
    \vspace{0.05em} % Changed from 1em to 0.5em for smaller spacing

    % --- Bottom Row (Swapped) ---
    % Using [t] here as well for consistency
    \begin{subfigure}[t]{0.98\columnwidth}  % <-- CHANGED [b] to [t]
        \centering
        % trim={<left> <bottom> <right> <top>}
        % We are trimming 8pt from the top (our manually calculated 10%)
        \includegraphics[width=\linewidth, 
                         trim={0 0 0 120pt},  % <-- Manually set 10%
                         clip]
                        {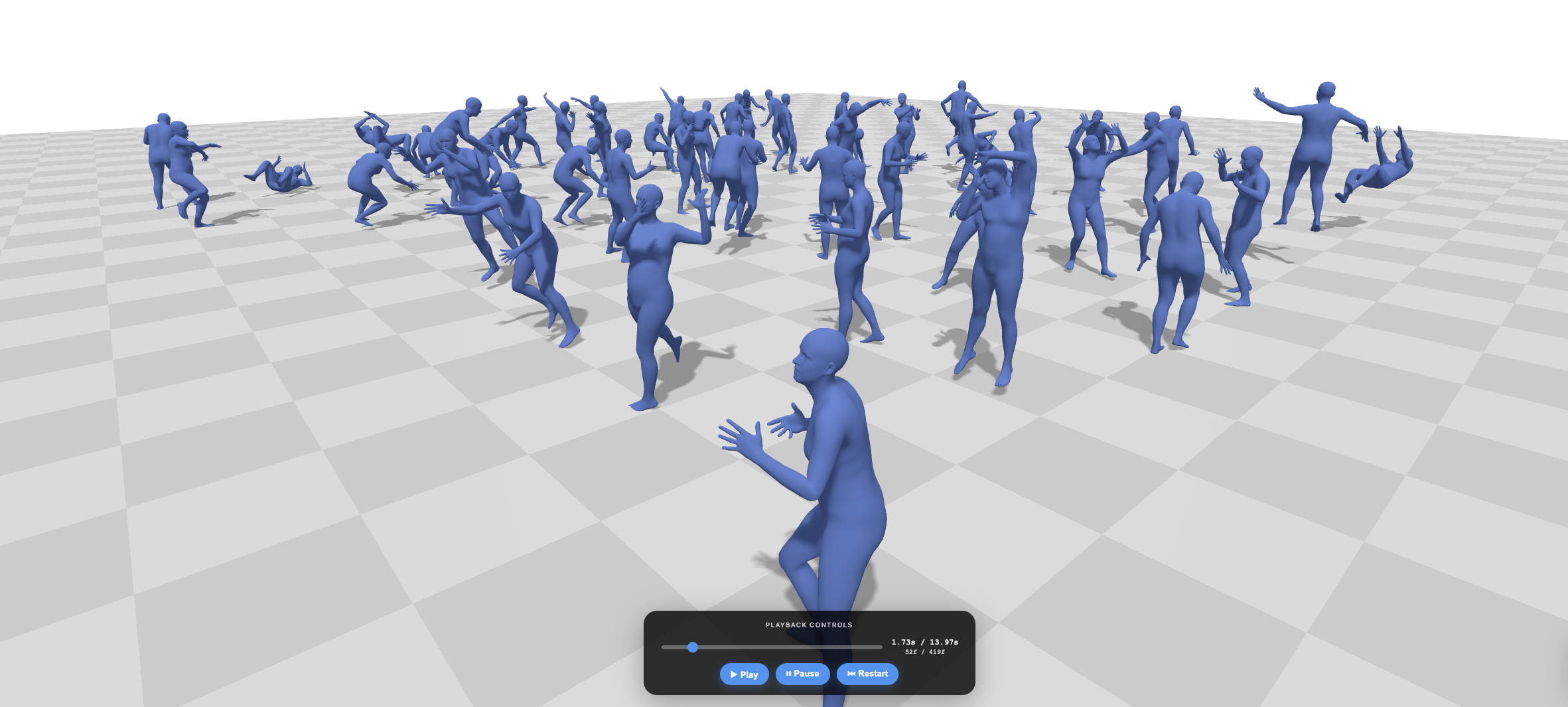}
        
        % \caption{} % Will be (c)
        \label{fig:bottom-center-new} % New label
    \end{subfigure}
    
    \end{minipage}
    % --- Overall Caption ---
    % Updated caption descriptions to match the new image order
    \caption{\textbf{Motion toolbox.} Our toolbox provides useful tools for visualizing keyframes (top left), a suite of evaluation metrics (top right) and an html multiple animations visuazlier (bottom).}
    \label{fig:toolbox_teaser} 
\end{figure}

We introduce the Motion Toolbox, a Python library designed to standardize 3D human motion analysis and quality assessment.
It provides a unified framework supporting multiple representations, including MotionMillion, HumanML3D, and various SMPL formats, with built-in converters for broad interoperability. 
The toolbox integrates research-validated metrics for physical plausibility—such as foot skating, ground penetration, jerk, and floating, proposed in prior works \cite{tevet2025closd, ron2025hoidini, yuan2023physdiff, mu2025stablemotion}.
Additionally, it features a web browser-based visualizer, enabling interactive inspection and rigorous evaluation of motion generation models.
\figref{fig:toolbox_teaser} Shows a subset of functionalities available in the toolbox, including the HTML motion visualizer and evaluation metric star chart. Please see HTML files in supplemental material to see the visualizer in action.

%%%%%%%%%%%%%%%%%%%%%%%%%%%%%%%%%%
%%%%%%%%%%% Experiment %%%%%%%%%%%
%%%%%%%%%%%%%%%%%%%%%%%%%%%%%%%%%%

\section{Experiments}
\label{sec:experiment}
\begin{figure}
    \centering
    \includegraphics[width=0.8\linewidth]{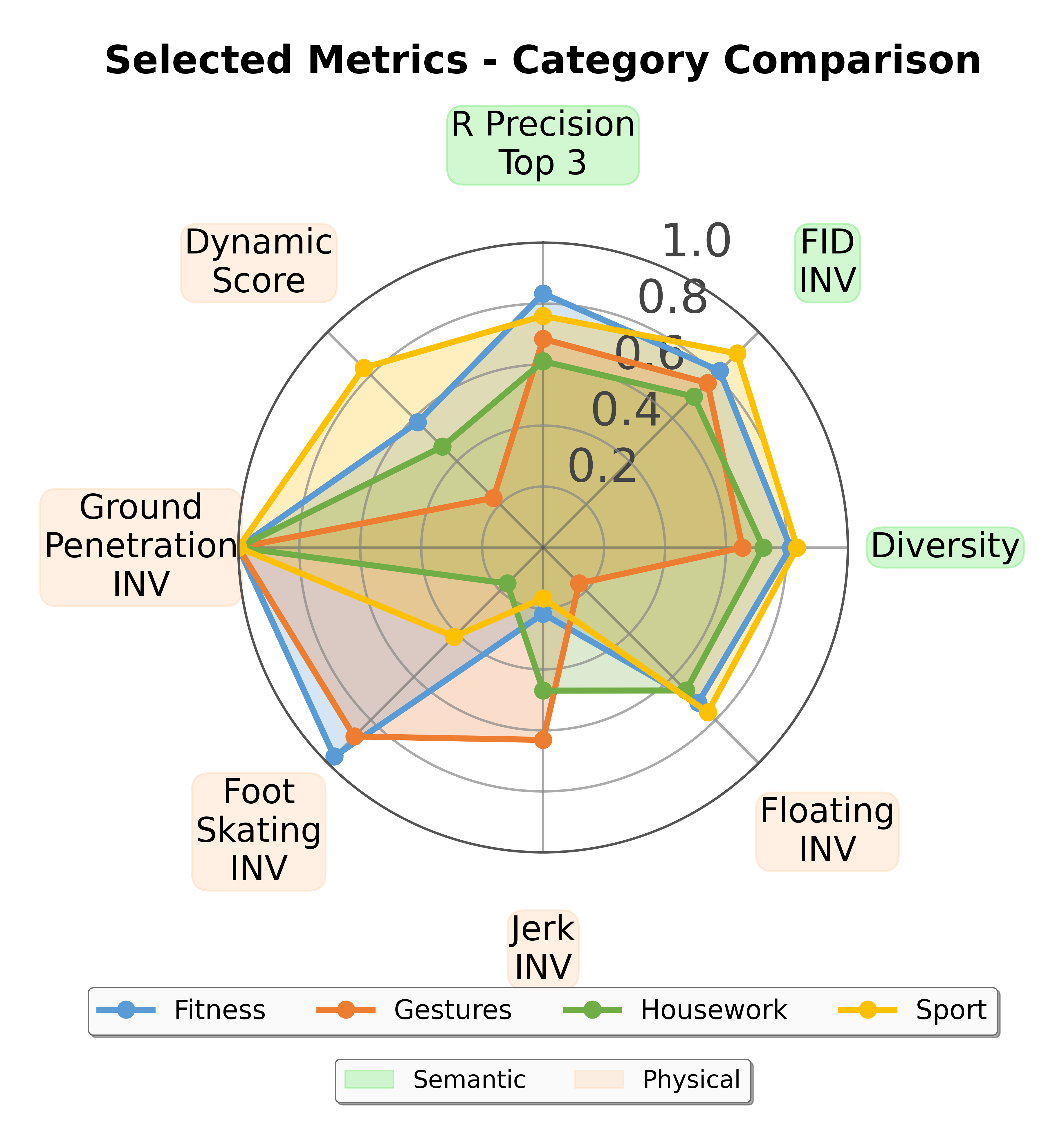}
    \caption{\textbf{Category wise evaluation of MDM~\cite{tevet2023human} on RoMo.} We report the evaluation metrics on different categories and how the significant differnces between them, highlighting the blind spots of the traditional aggregated reporting method.
    Some metrics were inverted (INV) to indicate higher-is-better.}
    \label{fig:mdm_category}
\end{figure}

\begin{table*}[t]
\centering
\small
\caption{Motion generation performance for MDM (diffusion) and MMGPT (GPT) models on RoMo.}
\setlength{\tabcolsep}{4pt}
\begin{tabular}{lccccccc}
\toprule
\textbf{\makecell{Method}} & \textbf{\makecell{Diversity $\uparrow$}} & \textbf{\makecell{FID $\downarrow$}} & \textbf{\makecell{Matching \\ Score $\uparrow$}} & \textbf{\makecell{Dynamic \\ Score $\uparrow$}} & \textbf{\makecell{Ground \\ Penetration $\downarrow$ \\ ($\times 10^{-5}$)}} & \textbf{\makecell{Foot \\ Skating $\downarrow$ \\ ($\times 10^{-3}$)}} & \textbf{\makecell{Floating $\downarrow$ \\ ($\times 10^{-2}$)}} \\
\midrule
MDM & 27.67 & 20.63 & 12.06 & 0.2138 & 0.0 & 1.70 & 1.67 \\
MMGPT & 16.68 & 12.80 & 22.08 & 0.3268 & 3.55 & 92.0 & 0.0311 \\
\bottomrule
\end{tabular}
\label{tab:sota_comparison}
\end{table*}

Our evaluation framework follows standard practices in the field by assessing two key aspects of motion generation: 
(1) semantic alignment with the input text and 
(2) the physical fidelity of the resulting motion.
For semantic alignment, we adopt Diversity, FID, Matching score, Dynamic score from prior work \cite{tevet2023human, guo2024momask, huang2024stablemofusion}.
An evaluator was trained exclusively for text-motion alignment, based on the TMR~\cite{petrovich2023tmr} framework. 
This framework employs a reduced dimension that contains only root translation and rotation information as well as the joint rotations \cite{hwangsnapmogen, guo2024momask}.
Additionally, we evaluate physical fidelity. This is achieved by employing an established suite of physical metrics from our toolbox designed to detect common artifacts, including foot skating, ground penetration, floating, and jerk/jitter \cite{karunratanakul2023gmd, ron2025hoidini, yuan2023physdiff, mu2025stablemotion}. 

\textbf{Implementation details.}
We train two models on RoMo: MDM \cite{tevet2023human}, a diffusion-based model and MotionMillionGPR (MMGPT) \cite{motionmillion} an autoregressive model. 
For MDM, a 50 diffusion step model was trained for 165k training steps using a transformer decoder architecture with a latent dimension of 512, feedforward size of 1024, 8 layers and 4 heads.
We used an Adam Optimizer with a learning rate of 1e-4 and batch size of 256.  A BERT text encoder was used to encode text up to a maximum length of 128. The MDM model was trained to output 224 frame animations. This was trained on a single A100-80GB GPU.

For MMGPT we trained as described in their work~\cite{motionmillion} with default settings. The VAE was trained with a batch size of 2048, while the 3B Llama GPT model was trained with a batch size of 16.
This model was trained on 4 A100-80GB GPUs, with 100k training iterations for the VAE and 75k training iterations for the GPT model.

\textbf{SOTA Model training.}
The results in \tabref{tab:sota_comparison} highlight a key architectural trade-off in motion generation.
MMGPT, an autoregressive model, excels in FID and Matching Score. This suggests its token-based, sequential approach is highly effective at capturing the precise semantic mapping from text to motion (Matching Score) and modeling the fine-grained statistical properties of the real motion distribution (FID).
Conversely, the diffusion-based MDM achieves superior Diversity and physical performance. 
Its denoising mechanism naturally produces a more varied set of motions for a single prompt. More importantly, its holistic, non-autoregressive refinement of the entire sequence appears to prevent the error accumulation common in sequential models, leading to motions with greater long-range consistency and physical plausibility.
Interestingly, the GPT model achieved a mean dynamic score closer to the RoMo ground truth than MDM, highlighting its temporal effectiveness.

\textbf{Taxonomy uncovers blind spots.} We evaluate the performance of the MDM model on subcategories from our taxonomy and report the performance  in \figref{fig:mdm_category}. The results show that different categories portray vastly different performance in multiple metrics, highlighting the blind sports that exist when aggregated evaluation is performed. 

%%%%%%%%%%%%%%%%%%%%%%%%%%%%%%%%%%
%%%%%%%%%%% Conclusion %%%%%%%%%%%
%%%%%%%%%%%%%%%%%%%%%%%%%%%%%%%%%%

\section{Conclusions and Future Work}
\label{sec:conclusion}

We introduced \textbf{\OurDatasetName{}} to address the critical lack of high-fidelity, large-scale human motion data. By employing adaptive filtering on in-the-wild sequences, our dataset successfully bridges the gap between constrained motion capture and noisy web collections. Beyond data, our novel \textbf{hierarchical taxonomy} transforms evaluation from opaque global metrics to transparent, per-category analysis.

Supported by the \textbf{Motion Toolbox} for standardized analysis, \OurDatasetName{} establishes a robust baseline for reproducible research and the next generation of truly generalizable human motion models.
Our taxonomy-guided experiments demonstrate how fine-grained evaluation sheds new light on where generative models succeed and where they fall short, offering clarity that previous datasets could not provide. We encourage the community to leverage this framework to advance the next wave of high-fidelity, broadly capable human motion generation models.

%%%%%%%%%%%%%%%%%%%%%%%%%%%%%%%%%
%%%%%%%%%%% Reference %%%%%%%%%%%
%%%%%%%%%%%%%%%%%%%%%%%%%%%%%%%%%
\clearpage
{
    \small
    \bibliographystyle{ieeenat_fullname}
    \bibliography{main}

@string(CVPR= {IEEE Conf. Comput. Vis. Pattern Recog.})

@string(ICCV= {Int. Conf. Comput. Vis.})

@string(ECCV= {Eur. Conf. Comput. Vis.})

@string(BMVC= {Brit. Mach. Vis. Conf.})

@string(TOG= {ACM Trans. Graph.})

@string(CVPR  = {CVPR})

@string(ICCV  = {ICCV})

@string(ECCV  = {ECCV})

@string(BMVC  =	{BMVC})

@string(TOG   = {ACM TOG})

@inproceedings{shin2024wham,
  title = {Wham: Reconstructing world-grounded humans with accurate 3d motion},
  author = {Shin, Soyong and Kim, Juyong and Halilaj, Eni and Black, Michael J},
  year = 2024,
  booktitle = {Proceedings of the IEEE/CVF Conference on Computer Vision and Pattern Recognition},
  pages = {2070--2080}
}

@article{motionlib,
  title = {Scaling Large Motion Models with Million-Level Human Motions},
  author = {Wang, Ye and Zheng, Sipeng and Cao, Bin and Wei, Qianshan and Zeng, Weishuai and Jin, Qin and Lu, Zongqing},
  year = 2024,
  journal = {arXiv preprint arXiv:2410.03311}
}

@article{motionx,
  title = {Motion-x: A large-scale 3d expressive whole-body human motion dataset},
  author = {Lin, Jing and Zeng, Ailing and Lu, Shunlin and Cai, Yuanhao and Zhang, Ruimao and Wang, Haoqian and Zhang, Lei},
  year = 2023,
  journal = {Advances in Neural Information Processing Systems},
  volume = 36,
  pages = {25268--25280}
}

@article{zhang2025motionx++,
  title = {Motion-x++: A large-scale multimodal 3d whole-body human motion dataset},
  author = {Zhang, Yuhong and Lin, Jing and Zeng, Ailing and Wu, Guanlin and Lu, Shunlin and Fu, Yurong and Cai, Yuanhao and Zhang, Ruimao and Wang, Haoqian and Zhang, Lei},
  year = 2025,
  journal = {arXiv preprint arXiv:2501.05098}
}

@inproceedings{motionmillion,
  title = {Go to zero: Towards zero-shot motion generation with million-scale data},
  author = {Fan, Ke and Lu, Shunlin and Dai, Minyue and Yu, Runyi and Xiao, Lixing and Dou, Zhiyang and Dong, Junting and Ma, Lizhuang and Wang, Jingbo},
  year = 2025,
  booktitle = {Proceedings of the IEEE/CVF International Conference on Computer Vision},
  pages = {13336--13348}
}

@inproceedings{hwangsnapmogen,
  title = {SnapMoGen: Human Motion Generation from Expressive Texts},
  author = {Hwang, Inwoo and Wang, Jian and Zhou, Bing and others},
  booktitle = {The Thirty-ninth Annual Conference on Neural Information Processing Systems}
}

@inproceedings{petrovich2023tmr,
  title = {Tmr: Text-to-motion retrieval using contrastive 3d human motion synthesis},
  author = {Petrovich, Mathis and Black, Michael J and Varol, G{\"u}l},
  year = 2023,
  booktitle = {Proceedings of the IEEE/CVF International Conference on Computer Vision},
  pages = {9488--9497}
}

@inproceedings{punnakkal2021babel,
  title = {BABEL: Bodies, action and behavior with english labels},
  author = {Punnakkal, Abhinanda R and Chandrasekaran, Arjun and Athanasiou, Nikos and Quiros-Ramirez, Alejandra and Black, Michael J},
  year = 2021,
  booktitle = {Proceedings of the IEEE/CVF conference on computer vision and pattern recognition},
  pages = {722--731}
}

@article{radford2019language,
  title = {Language Models are Unsupervised Multitask Learners},
  author = {Radford, Alec and Wu, Jeff and Child, Rewon and Luan, David and Amodei, Dario and Sutskever, Ilya},
  year = 2019
}

@article{lee2019dancing,
  title = {Dancing to music},
  author = {Lee, Hsin-Ying and Yang, Xiaodong and Liu, Ming-Yu and Wang, Ting-Chun and Lu, Yu-Ding and Yang, Ming-Hsuan and Kautz, Jan},
  year = 2019,
  journal = {Advances in neural information processing systems},
  volume = 32
}

@inproceedings{siyao2022bailando,
  title = {Bailando: 3d dance generation by actor-critic gpt with choreographic memory},
  author = {Siyao, Li and Yu, Weijiang and Gu, Tianpei and Lin, Chunze and Wang, Quan and Qian, Chen and Loy, Chen Change and Liu, Ziwei},
  year = 2022,
  booktitle = {Proceedings of the IEEE/CVF Conference on Computer Vision and Pattern Recognition},
  pages = {11050--11059}
}

@inproceedings{zhong2024smoodi,
  title = {SMooDi: Stylized Motion Diffusion Model},
  author = {Zhong, Lei and Xie, Yiming and Jampani, Varun and Sun, Deqing and Jiang, Huaizu},
  year = 2024,
  booktitle = {ECCV}
}

@misc{yolov8_ultralytics,
  title = {Ultralytics YOLOv8},
  author = {Glenn Jocher and Ayush Chaurasia and Jing Qiu},
  year = 2023,
  url = {https://github.com/ultralytics/ultralytics},
  version = {8.0.0},
  orcid = {0000-0001-5950-6979, 0000-0002-7603-6750, 0000-0003-3783-7069},
  license = {AGPL-3.0}
}

@article{qwen3,
  title = {Qwen3 Technical Report},
  author = {An Yang and Anfeng Li and Baosong Yang and Beichen Zhang and Binyuan Hui and Bo Zheng and Bowen Yu and Chang Gao and Chengen Huang and Chenxu Lv and Chujie Zheng and Dayiheng Liu and Fan Zhou and Fei Huang and Feng Hu and Hao Ge and Haoran Wei and Huan Lin and Jialong Tang and Jian Yang and Jianhong Tu and Jianwei Zhang and Jianxin Yang and Jiaxi Yang and Jing Zhou and Jingren Zhou and Junyang Lin and Kai Dang and Keqin Bao and Kexin Yang and Le Yu and Lianghao Deng and Mei Li and Mingfeng Xue and Mingze Li and Pei Zhang and Peng Wang and Qin Zhu and Rui Men and Ruize Gao and Shixuan Liu and Shuang Luo and Tianhao Li and Tianyi Tang and Wenbiao Yin and Xingzhang Ren and Xinyu Wang and Xinyu Zhang and Xuancheng Ren and Yang Fan and Yang Su and Yichang Zhang and Yinger Zhang and Yu Wan and Yuqiong Liu and Zekun Wang and Zeyu Cui and Zhenru Zhang and Zhipeng Zhou and Zihan Qiu},
  year = 2025,
  journal = {arXiv preprint arXiv:2505.09388}
}

@article{ron2025hoidini,
  title = {HOIDiNi: Human-Object Interaction through Diffusion Noise Optimization},
  author = {Ron, Roey and Tevet, Guy and Sawdayee, Haim and Bermano, Amit H},
  year = 2025,
  journal = {arXiv preprint arXiv:2506.15625}
}

@article{zhang2025bimart,
  title = {BimArt: A Unified Approach for the Synthesis of 3D Bimanual Interaction with Articulated Objects},
  author = {Zhang, Wanyue and Dabral, Rishabh and Golyanik, Vladislav and Choutas, Vasileios and Alvarado, Eduardo and Beeler, Thabo and Habermann, Marc and Theobalt, Christian},
  year = 2025,
  journal = {Proceedings of the IEEE/CVF Conference on Computer Vision and Pattern Recognition (CVPR)}
}

@inproceedings{guo_2022_cvpr,
  title = {Generating Diverse and Natural 3D Human Motions From Text},
  author = {Guo, Chuan and Zou, Shihao and Zuo, Xinxin and Wang, Sen and Ji, Wei and Li, Xingyu and Cheng, Li},
  year = 2022,
  month = {June},
  booktitle = {Proceedings of the IEEE/CVF Conference on Computer Vision and Pattern Recognition (CVPR)},
  pages = {5152--5161}
}

@inproceedings{tevet2023human,
  title = {Human Motion Diffusion Model},
  author = {Guy Tevet and Sigal Raab and Brian Gordon and Yoni Shafir and Daniel Cohen-or and Amit Haim Bermano},
  year = 2023,
  booktitle = {The Eleventh International Conference on Learning Representations},
  url = {https://openreview.net/forum?id=SJ1kSyO2jwu}
}

@inproceedings{dabral2023mofusion,
  title = {Mofusion: A framework for denoising-diffusion-based motion synthesis},
  author = {Dabral, Rishabh and Mughal, Muhammad Hamza and Golyanik, Vladislav and Theobalt, Christian},
  year = 2023,
  booktitle = {Proceedings of the IEEE/CVF Conference on Computer Vision and Pattern Recognition},
  pages = {9760--9770}
}

@inproceedings{shafir2024human,
  title = {Human Motion Diffusion as a Generative Prior},
  author = {Yoni Shafir and Guy Tevet and Roy Kapon and Amit Haim Bermano},
  year = 2024,
  booktitle = {The Twelfth International Conference on Learning Representations},
  url = {https://openreview.net/forum?id=dTpbEdN9kr}
}

@article{raab2024monkey,
  title = {Monkey See, Monkey Do: Harnessing Self-attention in Motion Diffusion for Zero-shot Motion Transfer},
  author = {Raab, Sigal and Gat, Inbar and Sala, Nathan and Tevet, Guy and Shalev-Arkushin, Rotem and Fried, Ohad and Bermano, Amit H and Cohen-Or, Daniel},
  year = 2024,
  journal = {arXiv preprint arXiv:2406.06508}
}

@article{karunratanakul2023gmd,
  title = {GMD: Controllable Human Motion Synthesis via Guided Diffusion Models},
  author = {Karunratanakul, Korrawe and Preechakul, Konpat and Suwajanakorn, Supasorn and Tang, Siyu},
  year = 2023,
  journal = {arXiv preprint arXiv:2305.12577}
}

@article{smpl:2015,
  title = {{SMPL}: A Skinned Multi-Person Linear Model},
  author = {Loper, Matthew and Mahmood, Naureen and Romero, Javier and Pons-Moll, Gerard and Black, Michael J.},
  year = 2015,
  month = oct,
  journal = {ACM Trans. Graphics (Proc. SIGGRAPH Asia)},
  publisher = {ACM},
  volume = 34,
  number = 6,
  pages = {248:1--248:16}
}

@article{xu2022vitpose,
  title = {Vitpose: Simple vision transformer baselines for human pose estimation},
  author = {Xu, Yufei and Zhang, Jing and Zhang, Qiming and Tao, Dacheng},
  year = 2022,
  journal = {Advances in Neural Information Processing Systems},
  volume = 35,
  pages = {38571--38584}
}

@inproceedings{tseng2023edge,
  title = {Edge: Editable dance generation from music},
  author = {Tseng, Jonathan and Castellon, Rodrigo and Liu, Karen},
  year = 2023,
  booktitle = {Proceedings of the IEEE/CVF Conference on Computer Vision and Pattern Recognition},
  pages = {448--458}
}

@inproceedings{karunratanakul2024optimizing,
  title = {Optimizing diffusion noise can serve as universal motion priors},
  author = {Karunratanakul, Korrawe and Preechakul, Konpat and Aksan, Emre and Beeler, Thabo and Suwajanakorn, Supasorn and Tang, Siyu},
  year = 2024,
  booktitle = {Proceedings of the IEEE/CVF Conference on Computer Vision and Pattern Recognition},
  pages = {1334--1345}
}

@inproceedings{yuan2023physdiff,
  title = {PhysDiff: Physics-Guided Human Motion Diffusion Model},
  author = {Yuan, Ye and Song, Jiaming and Iqbal, Umar and Vahdat, Arash and Kautz, Jan},
  year = 2023,
  booktitle = {Proceedings of the IEEE/CVF International Conference on Computer Vision (ICCV)}
}

@article{mahmood2019amassao,
  title = {AMASS: Archive of Motion Capture As Surface Shapes},
  author = {Naureen Mahmood and N. Ghorbani and N. Troje and Gerard Pons-Moll and Michael J. Black},
  year = 2019,
  journal = {2019 IEEE/CVF International Conference on Computer Vision (ICCV)},
  pages = {5441--5450}
}

@inproceedings{guo2024momask,
  title = {Momask: Generative masked modeling of 3d human motions},
  author = {Guo, Chuan and Mu, Yuxuan and Javed, Muhammad Gohar and Wang, Sen and Cheng, Li},
  year = 2024,
  booktitle = {Proceedings of the IEEE/CVF Conference on Computer Vision and Pattern Recognition},
  pages = {1900--1910}
}

@inproceedings{tang2019coin,
  title = {Coin: A large-scale dataset for comprehensive instructional video analysis},
  author = {Tang, Yansong and Ding, Dajun and Rao, Yongming and Zheng, Yu and Zhang, Danyang and Zhao, Lili and Lu, Jiwen and Zhou, Jie},
  year = 2019,
  booktitle = {Proceedings of the IEEE/CVF Conference on Computer Vision and Pattern Recognition},
  pages = {1207--1216}
}

@article{nan2024openvid,
  title = {OpenVid-1M: A Large-Scale High-Quality Dataset for Text-to-video Generation},
  author = {Nan, Kepan and Xie, Rui and Zhou, Penghao and Fan, Tiehan and Yang, Zhenheng and Chen, Zhijie and Li, Xiang and Yang, Jian and Tai, Ying},
  year = 2024,
  journal = {arXiv preprint arXiv:2407.02371}
}

@inproceedings{caba2015activitynet,
  title = {Activitynet: A large-scale video benchmark for human activity understanding},
  author = {Caba Heilbron, Fabian and Escorcia, Victor and Ghanem, Bernard and Carlos Niebles, Juan},
  year = 2015,
  booktitle = {Proceedings of the ieee conference on computer vision and pattern recognition},
  pages = {961--970}
}

@inproceedings{ben2021ikea,
  title = {The ikea asm dataset: Understanding people assembling furniture through actions, objects and pose},
  author = {Ben-Shabat, Yizhak and Yu, Xin and Saleh, Fatemeh and Campbell, Dylan and Rodriguez-Opazo, Cristian and Li, Hongdong and Gould, Stephen},
  year = 2021,
  booktitle = {Proceedings of the IEEE/CVF Winter Conference on Applications of Computer Vision},
  pages = {847--859}
}

@inproceedings{taheri2020grab,
  title = {GRAB: A dataset of whole-body human grasping of objects},
  author = {Taheri, Omid and Ghorbani, Nima and Black, Michael J and Tzionas, Dimitrios},
  year = 2020,
  booktitle = {Computer Vision--ECCV 2020: 16th European Conference, Glasgow, UK, August 23--28, 2020, Proceedings, Part IV 16},
  pages = {581--600},
  organization = {Springer}
}

@inproceedings{camdm,
  title = {Taming Diffusion Probabilistic Models for Character Control},
  author = {Chen, Rui and Shi, Mingyi and Huang, Shaoli and Tan, Ping and Komura, Taku and Chen, Xuelin},
  year = 2024,
  booktitle = {ACM SIGGRAPH 2024 Conference Papers},
  location = {Denver, CO, USA},
  publisher = {Association for Computing Machinery},
  address = {New York, NY, USA},
  series = {SIGGRAPH '24},
  doi = {10.1145/3641519.3657440},
  url = {https://doi.org/10.1145/3641519.3657440}
}

@article{plappert2016kit,
  title = {The KIT motion-language dataset},
  author = {Plappert, Matthias and Mandery, Christian and Asfour, Tamim},
  year = 2016,
  journal = {Big data},
  publisher = {Mary Ann Liebert, Inc. 140 Huguenot Street, 3rd Floor New Rochelle, NY 10801 USA},
  volume = 4,
  number = 4,
  pages = {236--252}
}

@inproceedings{guo2022generating,
  title = {Generating Diverse and Natural 3D Human Motions From Text},
  author = {Guo, Chuan and Zou, Shihao and Zuo, Xinxin and Wang, Sen and Ji, Wei and Li, Xingyu and Cheng, Li},
  year = 2022,
  booktitle = {Proceedings of the IEEE/CVF Conference on Computer Vision and Pattern Recognition},
  publisher = {IEEE Computer Society},
  address = {Washington, DC, USA},
  pages = {5152--5161}
}

@article{ho2020denoising,
  title = {Denoising diffusion probabilistic models},
  author = {Ho, Jonathan and Jain, Ajay and Abbeel, Pieter},
  year = 2020,
  journal = {Advances in Neural Information Processing Systems},
  volume = 33,
  pages = {6840--6851}
}

@inproceedings{tevet2022motionclip,
  title = {Motionclip: Exposing human motion generation to clip space},
  author = {Tevet, Guy and Gordon, Brian and Hertz, Amir and Bermano, Amit H and Cohen-Or, Daniel},
  year = 2022,
  booktitle = {Computer Vision--ECCV 2022: 17th European Conference, Tel Aviv, Israel, October 23--27, 2022, Proceedings, Part XXII},
  publisher = {Springer International Publishing},
  address = {Berlin/Heidelberg, Germany},
  pages = {358--374},
  organization = {Springer}
}

@inproceedings{petrovich2021actor,
  title = {Action-Conditioned 3{D} Human Motion Synthesis with Transformer {VAE}},
  author = {Petrovich, Mathis and Black, Michael J. and Varol, G{\"u}l},
  year = 2021,
  month = {October},
  booktitle = {International Conference on Computer Vision (ICCV)},
  publisher = {IEEE Computer Society},
  address = {Washington, DC, USA},
  pages = {10985--10995}
}

@article{vaswani2017attention,
  title = {Attention is all you need},
  author = {Vaswani, Ashish and Shazeer, Noam and Parmar, Niki and Uszkoreit, Jakob and Jones, Llion and Gomez, Aidan N and Kaiser, {\L}ukasz and Polosukhin, Illia},
  year = 2017,
  journal = {Advances in neural information processing systems},
  volume = 30,
  pages = {}
}

@article{meng2024rethinking,
  title = {Rethinking Diffusion for Text-Driven Human Motion Generation},
  author = {Meng, Zichong and Xie, Yiming and Peng, Xiaogang and Han, Zeyu and Jiang, Huaizu},
  year = 2024,
  journal = {arXiv preprint arXiv:2411.16575}
}

@article{heusel2017gans,
  title = {Gans trained by a two time-scale update rule converge to a local nash equilibrium},
  author = {Heusel, Martin and Ramsauer, Hubert and Unterthiner, Thomas and Nessler, Bernhard and Hochreiter, Sepp},
  year = 2017,
  journal = {Advances in neural information processing systems},
  volume = 30,
  pages = {}
}

@inproceedings{zhang2023generating,
  title = {T2M-GPT: Generating Human Motion from Textual Descriptions with Discrete Representations},
  author = {Zhang, Jianrong and Zhang, Yangsong and Cun, Xiaodong and Huang, Shaoli and Zhang, Yong and Zhao, Hongwei and Lu, Hongtao and Shen, Xi},
  year = 2023,
  booktitle = {Proceedings of the IEEE/CVF Conference on Computer Vision and Pattern Recognition (CVPR)},
  publisher = {IEEE Computer Society},
  address = {Washington, DC, USA},
  pages = {}
}

@article{jiang2024motiongpt,
  title = {MotionGPT: Human Motion as a Foreign Language},
  author = {Jiang, Biao and Chen, Xin and Liu, Wen and Yu, Jingyi and Yu, Gang and Chen, Tao},
  year = 2024,
  journal = {Advances in Neural Information Processing Systems},
  volume = 36,
  pages = {}
}

@article{van2017neural,
  title = {Neural discrete representation learning},
  author = {Van Den Oord, Aaron and Vinyals, Oriol and others},
  year = 2017,
  journal = {Advances in neural information processing systems},
  volume = 30,
  pages = {}
}

@inproceedings{liang2024intergen,
  title = {Intergen: Diffusion-based multi-human motion generation under complex interactions},
  author = {Liang, Han and Zhang, Wenqian and Li, Wenxuan and Yu, Jingyi and Xu, Lan},
  year = 2024,
  booktitle = {International Journal of Computer Vision},
  publisher = {Springer},
  address = {Berlin/Heidelberg, Germany},
  pages = {1--21}
}

@misc{cohan2024flexible,
  title = {Flexible Motion In-betweening with Diffusion Models},
  author = {Cohan, Setareh and Tevet, Guy and Reda, Daniele and Peng, Xue Bin and van de Panne, Michiel},
  year = 2024,
  journal = {arXiv preprint arXiv:2405.11126}
}

@article{alexanderson2023listen,
  title = {Listen, denoise, action! audio-driven motion synthesis with diffusion models},
  author = {Alexanderson, Simon and Nagy, Rajmund and Beskow, Jonas and Henter, Gustav Eje},
  year = 2023,
  journal = {ACM Transactions on Graphics (TOG)},
  publisher = {ACM New York, NY, USA},
  volume = 42,
  number = 4,
  pages = {1--20}
}

@inproceedings{tevet2025closd,
  title = {{CL}o{SD}: Closing the Loop between Simulation and Diffusion for multi-task character control},
  author = {Guy Tevet and Sigal Raab and Setareh Cohan and Daniele Reda and Zhengyi Luo and Xue Bin Peng and Amit Haim Bermano and Michiel van de Panne},
  year = 2025,
  booktitle = {The Thirteenth International Conference on Learning Representations},
  url = {https://openreview.net/forum?id=pZISppZSTv}
}

@article{sawdayee2025dance,
  title = {Dance Like a Chicken: Low-Rank Stylization for Human Motion Diffusion},
  author = {Sawdayee, Haim and Guo, Chuan and Tevet, Guy and Zhou, Bing and Wang, Jian and Bermano, Amit H},
  year = 2025,
  journal = {arXiv preprint arXiv:2503.19557}
}

@inproceedings{li2024task,
  title = {Task-oriented human-object interactions generation with implicit neural representations},
  author = {Li, Quanzhou and Wang, Jingbo and Loy, Chen Change and Dai, Bo},
  year = 2024,
  booktitle = {Proceedings of the IEEE/CVF Winter Conference on Applications of Computer Vision},
  pages = {3035--3044}
}

@article{li2023object,
  title = {Object motion guided human motion synthesis},
  author = {Li, Jiaman and Wu, Jiajun and Liu, C Karen},
  year = 2023,
  journal = {ACM Transactions on Graphics (TOG)},
  publisher = {ACM New York, NY, USA},
  volume = 42,
  number = 6,
  pages = {1--11}
}

@article{peng2023hoi,
  title = {Hoi-diff: Text-driven synthesis of 3d human-object interactions using diffusion models},
  author = {Peng, Xiaogang and Xie, Yiming and Wu, Zizhao and Jampani, Varun and Sun, Deqing and Jiang, Huaizu},
  year = 2023,
  journal = {arXiv preprint arXiv:2312.06553}
}

@article{pi2025coda,
  title = {CoDA: Coordinated Diffusion Noise Optimization for Whole-Body Manipulation of Articulated Objects},
  author = {Pi, Huaijin and Cen, Zhi and Dou, Zhiyang and Komura, Taku},
  year = 2025,
  journal = {arXiv preprint arXiv:2505.21437}
}

@inproceedings{shen2024gvhmr,
  title = {World-Grounded Human Motion Recovery via Gravity-View Coordinates},
  author = {Shen, Zehong and Pi, Huaijin and Xia, Yan and Cen, Zhi and Peng, Sida and Hu, Zechen and Bao, Hujun and Hu, Ruizhen and Zhou, Xiaowei},
  year = 2024,
  booktitle = {SIGGRAPH Asia Conference Proceedings}
}

@article{mu2025stablemotion,
  title = {StableMotion: Training Motion Cleanup Models with Unpaired Corrupted Data},
  author = {Mu, Yuxuan and Ling, Hung Yu and Shi, Yi and Ojeda, Ismael Baira and Xi, Pengcheng and Shu, Chang and Zinno, Fabio and Peng, Xue Bin},
  year = 2025,
  journal = {arXiv preprint arXiv:2505.03154}
}

@inproceedings{joo2015panoptic,
  title = {Panoptic studio: A massively multiview system for social motion capture},
  author = {Joo, Hanbyul and Liu, Hao and Tan, Lei and Gui, Lin and Nabbe, Bart and Matthews, Iain and Kanade, Takeo and Nobuhara, Shohei and Sheikh, Yaser},
  year = 2015,
  booktitle = {Proceedings of the IEEE international conference on computer vision},
  pages = {3334--3342}
}

@article{ionescu2013human3,
  title = {Human3. 6m: Large scale datasets and predictive methods for 3d human sensing in natural environments},
  author = {Ionescu, Catalin and Papava, Dragos and Olaru, Vlad and Sminchisescu, Cristian},
  year = 2013,
  journal = {IEEE transactions on pattern analysis and machine intelligence},
  publisher = {IEEE},
  volume = 36,
  number = 7,
  pages = {1325--1339}
}

@inproceedings{mandery2015kit,
  title = {The KIT whole-body human motion database},
  author = {Mandery, Christian and Terlemez, {\"O}mer and Do, Martin and Vahrenkamp, Nikolaus and Asfour, Tamim},
  year = 2015,
  booktitle = {2015 International Conference on Advanced Robotics (ICAR)},
  pages = {329--336},
  organization = {IEEE}
}

@article{dip:siggraphasia:2018,
  title = {Deep Inertial Poser Learning to Reconstruct Human Pose from SparseInertial Measurements in Real Time},
  author = {Huang, Yinghao and Kaufmann, Manuel and Aksan, Emre and Black, Michael J. and Hilliges, Otmar and Pons-Moll, Gerard},
  year = 2018,
  month = nov,
  journal = {ACM Transactions on Graphics, (Proc. SIGGRAPH Asia)},
  volume = 37,
  number = 6,
  pages = {185:1--185:15}
}

@inproceedings{lu_2025_humoto,
  title = {HUMOTO: A 4D Dataset of Mocap Human Object Interactions},
  author = {Lu, Jiaxin and Huang, Chun-Hao Paul and Bhattacharya, Uttaran and Huang, Qixing and Zhou, Yi},
  year = 2025,
  month = {October},
  booktitle = {Proceedings of the IEEE/CVF International Conference on Computer Vision (ICCV)},
  pages = {10886--10897}
}

@inproceedings{tsuchida2019aist,
  title = {AIST Dance Video Database: Multi-Genre, Multi-Dancer, and Multi-Camera Database for Dance Information Processing.},
  author = {Tsuchida, Shuhei and Fukayama, Satoru and Hamasaki, Masahiro and Goto, Masataka},
  year = 2019,
  booktitle = {ISMIR},
  volume = 1,
  number = 5,
  pages = 6
}

@misc{amass_accad,
  title = {{ACCAD MoCap Dataset}},
  author = {{Advanced Computing Center for the Arts and Design}},
  url = {https://accad.osu.edu/research/motion-lab/mocap-system-and-data}
}

@inproceedings{amass_bmlhandball,
  title = {Bewegungsanalyse getäuschter und nicht-getäuschter 7m-Würfe im Handball},
  author = {Helm, Fabian and Troje, Nikolaus and Reiser, Mathias and Munzert, Jörn},
  year = 2015,
  month = {01},
  journal = {47. Jahrestagung der Arbeitsgemeinschaft für Sportpsychologie, Freiburg.},
  pages = {}
}

@article{amass_bmlmovi,
  title = {{MoVi}: A Large Multipurpose Motion and Video Dataset},
  author = {Saeed Ghorbani and Kimia Mahdaviani and Anne Thaler and Konrad Kording and Douglas James Cook and Gunnar Blohm and Nikolaus F. Troje},
  year = 2020,
  journal = {arXiv preprint arXiv: 2003.01888}
}

@article{amass_bmlrub,
  title = {Decomposing Biological Motion: {A} Framework for Analysis and Synthesis of Human Gait Patterns},
  author = {Troje, Nikolaus F.},
  year = 2002,
  month = sep,
  journal = {Journal of Vision},
  volume = 2,
  number = 5,
  pages = {2--2},
  doi = {10.1167/2.5.2},
  month_numeric = 9
}

@misc{amass_cmu,
  title = {{CMU MoCap Dataset}},
  author = {{Carnegie Mellon University}},
  url = {http://mocap.cs.cmu.edu}
}

@article{amass_dancedb,
  title = {Digital Dance Ethnography: {O}rganizing Large Dance Collections},
  author = {Aristidou, Andreas and Shamir, Ariel and Chrysanthou, Yiorgos},
  year = 2019,
  month = nov,
  journal = {J. Comput. Cult. Herit.},
  publisher = {Association for Computing Machinery},
  address = {New York, NY, USA},
  volume = 12,
  number = 4,
  doi = {10.1145/3344383},
  issn = {1556-4673},
  url = {https://doi.org/10.1145/3344383},
  issue_date = {January 2020},
  articleno = 29,
  numpages = 27,
  acmid = {}
}

@inproceedings{amass_dfaust,
  title = {Dynamic {FAUST}: {R}egistering Human Bodies in Motion},
  author = {Bogo, Federica and Romero, Javier and Pons-Moll, Gerard and Black, Michael J.},
  year = 2017,
  month = jul,
  booktitle = {IEEE Conf. on Computer Vision and Pattern Recognition (CVPR)},
  month_numeric = 7
}

@misc{amass_eyesjapandataset,
  title = {{Eyes Japan MoCap Dataset}},
  author = {Eyes JAPAN Co. Ltd.},
  url = {http://mocapdata.com}
}

@inproceedings{amass_grab,
  title = {{GRAB}: A Dataset of Whole-Body Human Grasping of Objects},
  author = {Taheri, Omid and Ghorbani, Nima and Black, Michael J. and Tzionas, Dimitrios},
  year = 2020,
  booktitle = {European Conference on Computer Vision (ECCV)},
  url = {https://grab.is.tue.mpg.de}
}

@inproceedings{amass_grab-2,
  title = {{ContactDB}: Analyzing and Predicting Grasp Contact via Thermal Imaging},
  author = {Brahmbhatt, Samarth and Ham, Cusuh and Kemp, Charles C. and Hays, James},
  year = 2019,
  booktitle = {The IEEE Conference on Computer Vision and Pattern Recognition (CVPR)},
  url = {https://contactdb.cc.gatech.edu}
}

@techreport{amass_hdm05,
  title = {Documentation Mocap Database HDM05},
  author = {M. M\"{u}ller and T. R\"{o}der and M. Clausen and B. Eberhardt and B. Kr\"{u}ger and A. Weber},
  year = 2007,
  month = {June},
  number = {CG-2007-2},
  issn = {1610-8892},
  institution = {Universit\"{a}t Bonn}
}

@article{amass_human4d,
  title = {HUMAN4D: A Human-Centric Multimodal Dataset for Motions and Immersive Media},
  author = {Chatzitofis, Anargyros and Saroglou, Leonidas and Boutis, Prodromos and Drakoulis, Petros and Zioulis, Nikolaos and Subramanyam, Shishir and Kevelham, Bart and Charbonnier, Caecilia and Cesar, Pablo and Zarpalas, Dimitrios and others},
  year = 2020,
  journal = {IEEE Access},
  publisher = {IEEE},
  volume = 8,
  pages = {176241--176262}
}

@article{amass_humaneva,
  title = {{HumanEva}: Synchronized video and motion capture dataset and baseline algorithm for evaluation of articulated human motion},
  author = {Sigal, L. and Balan, A. and Black, M. J.},
  year = 2010,
  month = mar,
  journal = {International Journal of Computer Vision},
  publisher = {Springer Netherlands},
  volume = 87,
  number = 1,
  pages = {4--27},
  doi = {},
  month_numeric = 3
}

@inproceedings{amass_kit-cnrs-ekut-weizmann,
  title = {The {KIT} Whole-Body Human Motion Database},
  author = {Christian Mandery and \"Omer Terlemez and Martin Do and Nikolaus Vahrenkamp and Tamim Asfour},
  year = 2015,
  booktitle = {International Conference on Advanced Robotics (ICAR)},
  pages = {329--336}
}

@article{amass_kit-cnrs-ekut-weizmann-2,
  title = {Unifying Representations and Large-Scale Whole-Body Motion Databases for Studying Human Motion},
  author = {Christian Mandery and \"Omer Terlemez and Martin Do and Nikolaus Vahrenkamp and Tamim Asfour},
  year = 2016,
  journal = {IEEE Transactions on Robotics},
  volume = 32,
  number = 4,
  pages = {796--809}
}

@inproceedings{amass_kit-cnrs-ekut-weizmann-3,
  title = {The {KIT} Bimanual Manipulation Dataset},
  author = {Franziska Krebs and Andre Meixner and Isabel Patzer and Tamim Asfour},
  year = 2021,
  booktitle = {IEEE/RAS International Conference on Humanoid Robots (Humanoids)},
  pages = {499--506}
}

@inproceedings{amass_moyo,
  title = {{3D} Human Pose Estimation via Intuitive Physics},
  author = {Tripathi, Shashank and M{\"u}ller, Lea and Huang, Chun-Hao P. and Taheri Omid and Black, Michael J. and Tzionas, Dimitrios},
  year = 2023,
  month = {June},
  booktitle = {Proceedings of the IEEE/CVF Conference on Computer Vision and Pattern Recognition (CVPR)}
}

@article{amass_mosh,
  title = {{MoSh}: Motion and Shape Capture from Sparse Markers},
  author = {Loper, Matthew M. and Mahmood, Naureen and Black, Michael J.},
  year = 2014,
  month = nov,
  journal = {ACM Transactions on Graphics, (Proc. SIGGRAPH Asia)},
  publisher = {ACM},
  address = {New York, NY, USA},
  volume = 33,
  number = 6,
  pages = {220:1--220:13},
  doi = {10.1145/2661229.2661273},
  url = {http://doi.acm.org/10.1145/2661229.2661273}
}

@inproceedings{amass_poseprior,
  title = {Pose-Conditioned Joint Angle Limits for {3D} Human Pose Reconstruction},
  author = {Akhter, Ijaz and Black, Michael J.},
  year = 2015,
  month = jun,
  booktitle = {IEEE Conf. on Computer Vision and Pattern Recognition (CVPR) 2015}
}

@misc{amass_sfu,
  title = {{SFU Motion Capture Database}},
  author = {Simon Fraser University and National University of Singapore},
  url = {http://mocap.cs.sfu.ca/}
}

@inproceedings{amass_soma,
  title = {{SOMA}: Solving Optical Marker-Based MoCap Automatically},
  author = {Ghorbani, Nima and Black, Michael J.},
  year = 2021,
  month = oct,
  booktitle = {Proc. International Conference on Computer Vision (ICCV)},
  pages = {11117--11126},
  doi = {},
  month_numeric = 10
}

@inproceedings{amass_tcdhands,
  title = {Sleight of Hand: Perception of Finger Motion from Reduced Marker Sets},
  author = {Ludovic Hoyet and Kenneth Ryall and Rachel McDonnell and Carol O'Sullivan},
  year = 2012,
  booktitle = {Proceedings of the ACM SIGGRAPH Symposium on Interactive 3D Graphics and Games},
  pages = {79--86},
  doi = {10.1145/2159616.2159629}
}

@inproceedings{amass_totalcapture,
  title = {{Total Capture}: 3D Human Pose Estimation Fusing Video and Inertial Sensors},
  author = {Trumble, Matt and Gilbert, Andrew and Malleson, Charles and Hilton, Adrian and Collomosse, John},
  year = 2017,
  booktitle = {2017 British Machine Vision Conference (BMVC)}
}

@inproceedings{amass_wheelposer,
  title = {WheelPoser: Sparse-IMU Based Body Pose Estimation for Wheelchair Users},
  author = {Li, Yunzhi and Mollyn, Vimal and Yuan, Kuang and Carrington, Patrick},
  year = 2024,
  booktitle = {Proceedings of the 26th International ACM SIGACCESS Conference on Computers and Accessibility},
  pages = {1--17}
}

@article{schuhmann2021laion,
  title = {Laion-400m: Open dataset of clip-filtered 400 million image-text pairs},
  author = {Schuhmann, Christoph and Vencu, Richard and Beaumont, Romain and Kaczmarczyk, Robert and Mullis, Clayton and Katta, Aarush and Coombes, Theo and Jitsev, Jenia and Komatsuzaki, Aran},
  year = 2021,
  journal = {arXiv preprint arXiv:2111.02114}
}

@inproceedings{ye2023decoupling,
  title = {Decoupling human and camera motion from videos in the wild},
  author = {Ye, Vickie and Pavlakos, Georgios and Malik, Jitendra and Kanazawa, Angjoo},
  year = 2023,
  booktitle = {Proceedings of the IEEE/CVF conference on computer vision and pattern recognition},
  pages = {21222--21232}
}

@inproceedings{muller2021self,
  title = {On self-contact and human pose},
  author = {Muller, Lea and Osman, Ahmed AA and Tang, Siyu and Huang, Chun-Hao P and Black, Michael J},
  year = 2021,
  booktitle = {Proceedings of the IEEE/CVF Conference on Computer Vision and Pattern Recognition},
  pages = {9990--9999}
}

@inproceedings{barquero2024flowmdm,
  title = {Seamless human motion composition with blended positional encodings},
  author = {Barquero, German and Escalera, Sergio and Palmero, Cristina},
  year = 2024,
  booktitle = {Proceedings of the IEEE/CVF Conference on Computer Vision and Pattern Recognition},
  pages = {457--469}
}

@inproceedings{carreira2017kinetics,
  title = {Quo vadis, action recognition? a new model and the kinetics dataset},
  author = {Carreira, Joao and Zisserman, Andrew},
  year = 2017,
  booktitle = {proceedings of the IEEE Conference on Computer Vision and Pattern Recognition},
  pages = {6299--6308}
}

@article{wang2025videoufo,
  title = {Videoufo: A million-scale user-focused dataset for text-to-video generation},
  author = {Wang, Wenhao and Yang, Yi},
  year = 2025,
  journal = {arXiv preprint arXiv:2503.01739}
}

@inproceedings{liu2025hoigen,
  title = {Hoigen-1m: A large-scale dataset for human-object interaction video generation},
  author = {Liu, Kun and Liu, Qi and Liu, Xinchen and Li, Jie and Zhang, Yongdong and Luo, Jiebo and He, Xiaodong and Liu, Wu},
  year = 2025,
  booktitle = {Proceedings of the Computer Vision and Pattern Recognition Conference},
  pages = {24001--24010}
}

@inproceedings{li2025openhumanvid,
  title = {Openhumanvid: A large-scale high-quality dataset for enhancing human-centric video generation},
  author = {Li, Hui and Xu, Mingwang and Zhan, Yun and Mu, Shan and Li, Jiaye and Cheng, Kaihui and Chen, Yuxuan and Chen, Tan and Ye, Mao and Wang, Jingdong and others},
  year = 2025,
  booktitle = {Proceedings of the Computer Vision and Pattern Recognition Conference},
  pages = {7752--7762}
}

@inproceedings{huang2024stablemofusion,
  title = {Stablemofusion: Towards robust and efficient diffusion-based motion generation framework},
  author = {Huang, Yiheng and Yang, Hui and Luo, Chuanchen and Wang, Yuxi and Xu, Shibiao and Zhang, Zhaoxiang and Zhang, Man and Peng, Junran},
  year = 2024,
  booktitle = {Proceedings of the 32nd ACM International Conference on Multimedia},
  pages = {224--232}
}

@misc{flux2024,
  title = {FLUX},
  author = {Black Forest Labs},
  year = 2024,
  howpublished = {\url{https://github.com/black-forest-labs/flux}}
}

@article{grattafiori2024llama,
  title = {The llama 3 herd of models},
  author = {Grattafiori, Aaron and Dubey, Abhimanyu and Jauhri, Abhinav and Pandey, Abhinav and Kadian, Abhishek and Al-Dahle, Ahmad and Letman, Aiesha and Mathur, Akhil and Schelten, Alan and Vaughan, Alex and others},
  year = 2024,
  journal = {arXiv preprint arXiv:2407.21783}
}

@article{wan2025,
  title = {Wan: Open and Advanced Large-Scale Video Generative Models},
  author = {Team Wan and Ang Wang and Baole Ai and Bin Wen and Chaojie Mao and Chen-Wei Xie and Di Chen and Feiwu Yu and Haiming Zhao and Jianxiao Yang and Jianyuan Zeng and Jiayu Wang and Jingfeng Zhang and Jingren Zhou and Jinkai Wang and Jixuan Chen and Kai Zhu and Kang Zhao and Keyu Yan and Lianghua Huang and Mengyang Feng and Ningyi Zhang and Pandeng Li and Pingyu Wu and Ruihang Chu and Ruili Feng and Shiwei Zhang and Siyang Sun and Tao Fang and Tianxing Wang and Tianyi Gui and Tingyu Weng and Tong Shen and Wei Lin and Wei Wang and Wei Wang and Wenmeng Zhou and Wente Wang and Wenting Shen and Wenyuan Yu and Xianzhong Shi and Xiaoming Huang and Xin Xu and Yan Kou and Yangyu Lv and Yifei Li and Yijing Liu and Yiming Wang and Yingya Zhang and Yitong Huang and Yong Li and You Wu and Yu Liu and Yulin Pan and Yun Zheng and Yuntao Hong and Yupeng Shi and Yutong Feng and Zeyinzi Jiang and Zhen Han and Zhi-Fan Wu and Ziyu Liu},
  year = 2025,
  journal = {arXiv preprint arXiv:2503.20314}
}
}

\end{document}

\maketitle

\begin{abstract}

This supplementary material provides extensive quantitative and interactive support for the claims presented in our main paper. 
We strongly encourage readers to engage with the accompanying interactive website and video, which offer a full demo of the \textbf{Motion Toolbox}, a Mindmap visualization of the novel semantic taxonomy, 3D rendered visualizations of motion samples from each category, and detailed analysis plots of our adaptive filtering approach. 
This document extends the model evaluation to include \textbf{MoMask++} and \textbf{MDM+BERT}, leveraging the taxonomy for a fine-grained, category-based analysis that reveals architecture-specific trade-offs.
Furthermore, we present an ablation study quantifying the optimal trade-off in motion dynamics using our Dynamic Score ($S_{Dynamic}$), demonstrating that a refined subset ($V_C$) significantly improves generative performance (FID of $17.97$ vs. $20.63$ for the full set). 
Finally, we provide full implementation details for reproducibility and demonstrate the power of our hierarchical taxonomy with the \textbf{Context-Aware Adaptive Filtering methodology}, which strategically preserves authentic, category-specific motions while eliminating quality artifacts.

\end{abstract}

\begin{figure*}
    \centering
    \includegraphics[width=\linewidth]{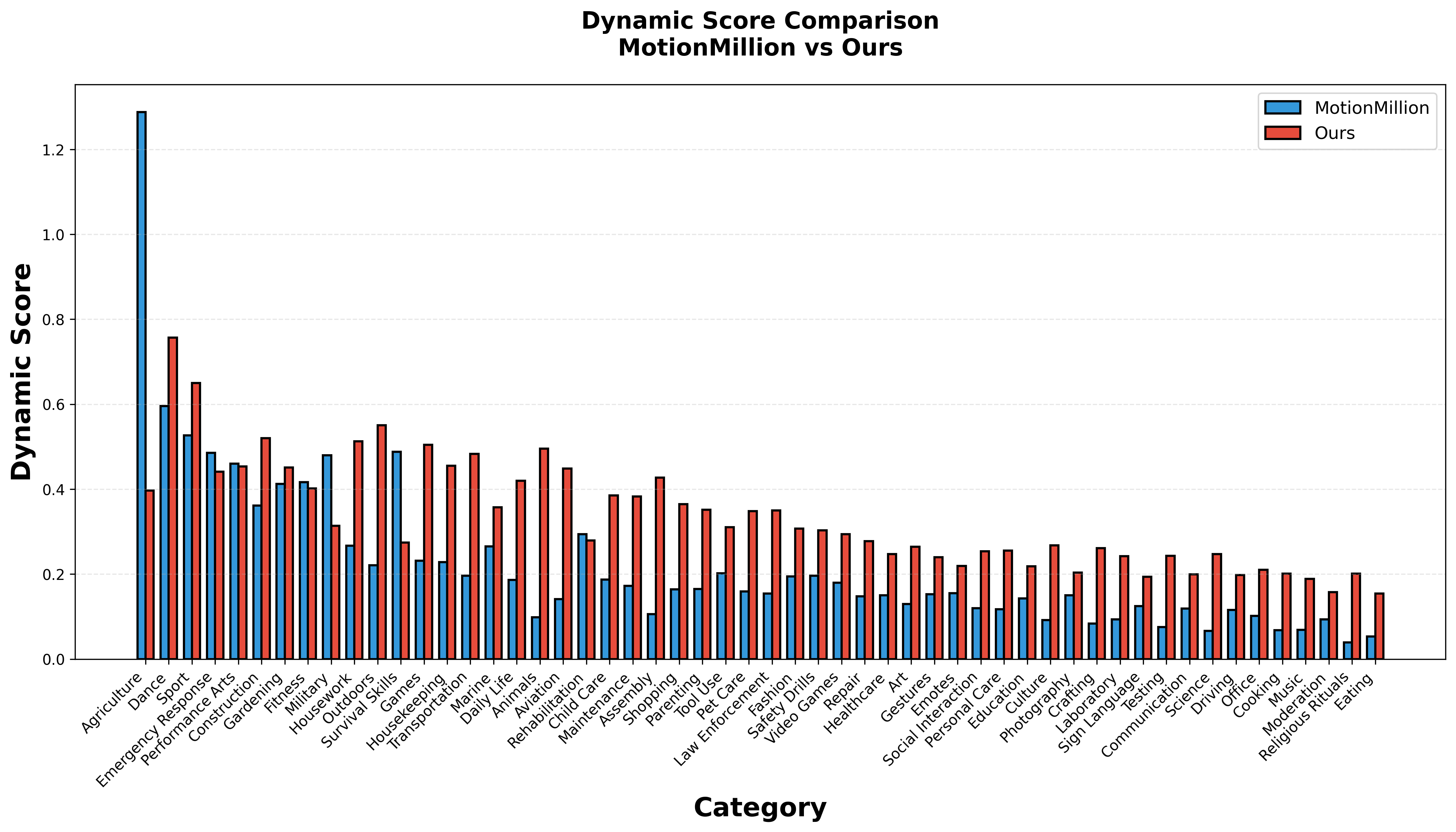}
    \caption{\textbf{Dynamic Score analysis.} \OurDatasetName{} demonstrates higher dynamic scores across the majority of categories}
    \label{fig:comparison_dynamic_score_sup}
\end{figure*}

\section{Taxonomy-Based Fine-Grained Evaluation}

Building upon the comparison of diffusion-based and GPT architectures in the main paper~(Sec.~6), we extend our evaluation to include \textbf{MoMask++}~\cite{hwangsnapmogen}  and \textbf{MDM+BERT}~\cite{tevet2023human}. We report the performance of these models in \tabref{tab:model_comparison_semantic} and \tabref{tab:model_comparison_physical} for semantic and physical metrics, respectively. Leveraging RoMo's broad taxonomy allows for a fine-grained analysis that surfaces intriguing trade-offs between these architectures. For example, while MDM demonstrates strong physical fidelity, MoMask++ achieves higher performance on semantic metrics, particularly FID (Fréchet Inception Distance), suggesting it better captures the statistical properties of the true motion distribution.

\begin{table*}[tb]
\centering
\caption{\textbf{Semantic benchmark.} Comparison of text-to-motion generation quality between MoMask++ and MDM trained on the RoMo full dataset.}
\begin{tabular}{lccccccc}
\toprule
\textbf{Method} & \multicolumn{3}{c}{\textbf{R Precision}} & \textbf{Diversity} $\uparrow$ & \textbf{FID} $\downarrow$ & \textbf{Matching Score} $\uparrow$ \\
& \textbf{Top1} & \textbf{Top2} & \textbf{Top3} & & & \\
\midrule
\textbf{MDM+BERT} & 0.5822 & 0.7578 & 0.8434 & 27.59 & 17.97 & 12.18 \\
\textbf{MoMask++} & 0.5148 & 0.6910 & 0.7820 & 28.17 & 14.30 & 12.86 \\
\bottomrule
\end{tabular}
\label{tab:model_comparison_semantic}
\end{table*}

\begin{table*}[tb]
\fontsize{9}{10}\selectfont
\centering
\caption{\textbf{Physical benchmark.} Comparison of physical motion artifacts and dynamic characteristics between MoMask++ and MDM.}
\begin{tabular}{lcccccc}
\toprule
\textbf{Method} & \textbf{Foot Skating} $\downarrow$ & \textbf{Jerk} $\downarrow$ & \textbf{Ground Penetration} $\downarrow$ & \textbf{Floating} $\uparrow$ & \textbf{Acceleration Peaks} $\uparrow$ & \textbf{Dynamic Score} $\uparrow$ \\
\midrule
\textbf{MDM+BERT} & 0.0011 & 46.77 & 0.00 & 0.0146 & 0.8374 & 0.2132 \\
\textbf{MoMask++} & 0.0022 & 332.73 & $2.1\mathrm{e}{-05}$ & 0.0020 & 4.7223 & 0.3887 \\
\bottomrule
\end{tabular}
\label{tab:model_comparison_physical}
\end{table*}

\begin{table*}[tb]
\fontsize{8.5}{9.5}\selectfont
\centering
\caption{\textbf{Taxonomy-based semantic evaluation.} Fine-grained breakdown of generation quality  comparing MDM and MoMask++ across the top 10 categories.}
\begin{tabular}{llcccccc}
\toprule
\textbf{Category} & \textbf{Method} & \multicolumn{3}{c}{\textbf{R Precision}} & \textbf{Diversity} $\uparrow$ & \textbf{FID} $\downarrow$ & \textbf{Matching Score} $\uparrow$ \\
& & \textbf{Top1} & \textbf{Top2} & \textbf{Top3} & & & \\
\midrule
Overall & MDM & 0.5822 & 0.7578 & 0.8434 & 27.59 & 17.97 & 12.18 \\
 & MoMask++ & 0.5148 & 0.6910 & 0.7820 & 28.17 & 14.30 & 12.86 \\
\midrule
Communication & MDM & 0.3771 & 0.5363 & 0.6393 & 21.60 & 28.97 & 9.77 \\
 & MoMask++ & 0.3152 & 0.4635 & 0.5621 & 23.13 & 14.64 & 11.08 \\
\midrule
Daily Life & MDM & 0.4782 & 0.6567 & 0.7537 & 26.55 & 27.80 & 12.15 \\
 & MoMask++ & 0.4066 & 0.5783 & 0.6760 & 26.56 & 21.27 & 12.91 \\
\midrule
Dance & MDM & 0.3514 & 0.5111 & 0.6105 & 22.88 & 32.02 & 12.56 \\
 & MoMask++ & 0.3088 & 0.4459 & 0.5344 & 21.48 & 37.09 & 12.67 \\
\midrule
Education & MDM & 0.3477 & 0.5221 & 0.6349 & 22.06 & 31.56 & 11.02 \\
 & MoMask++ & 0.3080 & 0.4641 & 0.5791 & 23.82 & 17.59 & 12.07 \\
\midrule
Fitness & MDM & 0.5777 & 0.7525 & 0.8350 & 27.35 & 23.78 & 13.01 \\
 & MoMask++ & 0.5098 & 0.6895 & 0.7764 & 26.05 & 19.36 & 12.99 \\
\midrule
Gestures & MDM & 0.4098 & 0.5723 & 0.6725 & 21.81 & 25.13 & 10.38 \\
 & MoMask++ & 0.3568 & 0.5125 & 0.6125 & 23.47 & 14.66 & 11.39 \\
\midrule
Housework & MDM & 0.3431 & 0.5093 & 0.6107 & 24.38 & 27.58 & 12.98 \\
 & MoMask++ & 0.3139 & 0.4832 & 0.5900 & 23.40 & 24.69 & 12.96 \\
\midrule
Outdoors & MDM & 0.4262 & 0.6125 & 0.7329 & 26.76 & 28.36 & 12.63 \\
 & MoMask++ & 0.3902 & 0.5689 & 0.6742 & 26.20 & 32.82 & 13.06 \\
\midrule
Pet Care & MDM & 0.3807 & 0.5834 & 0.7004 & 24.72 & 33.23 & 11.56 \\
 & MoMask++ & 0.3018 & 0.4846 & 0.6096 & 24.90 & 30.54 & 12.56 \\
\midrule
Sport & MDM & 0.4891 & 0.6820 & 0.7787 & 28.15 & 20.74 & 13.73 \\
 & MoMask++ & 0.4207 & 0.6158 & 0.7199 & 26.60 & 31.80 & 13.57 \\
\bottomrule
\end{tabular}
\label{tab:category_split_semantic}
\end{table*}

\begin{table*}[tb]
\fontsize{8.5}{9.5}\selectfont
\centering
\caption{\textbf{Taxonomy-based physical evaluation.} Fine-grained breakdown of physical motion quality and artifacts comparing MDM and MoMask++ across the top 10 categories.}
\begin{tabular}{llcccccc}
\toprule
\textbf{Category} & \textbf{Method} & \textbf{Foot Skating} $\downarrow$ & \textbf{Jerk} $\downarrow$ & \textbf{Ground Penetration} $\downarrow$ & \textbf{Floating} $\uparrow$ & \textbf{Accel. Peaks} $\uparrow$ & \textbf{Dynamic Score} $\uparrow$ \\
\midrule
Overall & MDM & 0.0011 & 46.77 & $0.0\mathrm{e}{+0}$ & 0.0146 & 0.8374 & 0.2132 \\
 & MoMask++ & 0.0022 & 332.73 & $2.1\mathrm{e}{-5}$ & 0.0020 & 4.7223 & 0.3887 \\
\midrule
Communication & MDM & 0.0009 & 32.51 & $0.0\mathrm{e}{+0}$ & 0.0217 & 0.4239 & 0.0959 \\
 & MoMask++ & 0.0027 & 217.11 & $1.8\mathrm{e}{-6}$ & 0.0043 & 2.5882 & 0.2010 \\
\midrule
Daily Life & MDM & 0.0036 & 49.18 & $0.0\mathrm{e}{+0}$ & 0.0054 & 1.0127 & 0.3941 \\
 & MoMask++ & 0.0019 & 320.67 & $5.1\mathrm{e}{-6}$ & 0.0010 & 4.6117 & 0.5264 \\
\midrule
Dance & MDM & 0.0004 & 72.88 & $0.0\mathrm{e}{+0}$ & 0.0037 & 1.9088 & 0.3031 \\
 & MoMask++ & 0.0005 & 510.93 & $6.4\mathrm{e}{-6}$ & 0.0004 & 8.3472 & 0.5850 \\
\midrule
Education & MDM & 0.0001 & 30.52 & $0.0\mathrm{e}{+0}$ & 0.0252 & 0.4201 & 0.1055 \\
 & MoMask++ & 0.0009 & 192.41 & $3.7\mathrm{e}{-7}$ & 0.0051 & 2.3672 & 0.1986 \\
\midrule
Fitness & MDM & 0.0000 & 60.89 & $0.0\mathrm{e}{+0}$ & 0.0073 & 1.1000 & 0.2447 \\
 & MoMask++ & 0.0003 & 432.76 & $1.6\mathrm{e}{-8}$ & 0.0008 & 6.1700 & 0.4781 \\
\midrule
Gestures & MDM & 0.0003 & 33.17 & $0.0\mathrm{e}{+0}$ & 0.0323 & 0.4644 & 0.1048 \\
 & MoMask++ & 0.0013 & 223.13 & $1.4\mathrm{e}{-6}$ & 0.0062 & 2.8128 & 0.2033 \\
\midrule
Housework & MDM & 0.0029 & 40.42 & $0.0\mathrm{e}{+0}$ & 0.0121 & 0.7514 & 0.1924 \\
 & MoMask++ & 0.0023 & 345.09 & $1.4\mathrm{e}{-6}$ & 0.0013 & 5.6696 & 0.4160 \\
\midrule
Outdoors & MDM & 0.0030 & 51.65 & $0.0\mathrm{e}{+0}$ & 0.0089 & 1.1816 & 0.4572 \\
 & MoMask++ & 0.0072 & 341.00 & $1.1\mathrm{e}{-5}$ & 0.0016 & 5.0046 & 0.5940 \\
\midrule
Pet Care & MDM & 0.0100 & 51.41 & $1.0\mathrm{e}{-4}$ & 0.0188 & 0.8560 & 0.3216 \\
 & MoMask++ & 0.0225 & 306.83 & $9.3\mathrm{e}{-6}$ & 0.0027 & 3.9456 & 0.4358 \\
\midrule
Sport & MDM & 0.0006 & 62.34 & $0.0\mathrm{e}{+0}$ & 0.0075 & 1.3625 & 0.3428 \\
 & MoMask++ & 0.0008 & 451.61 & $5.5\mathrm{e}{-6}$ & 0.0007 & 7.0758 & 0.6317 \\
\bottomrule
\end{tabular}
\label{tab:category_split_physical}
\end{table*}

\subsection{Taxonomy-Based Evaluation Insights}

Our hierarchical taxonomy exposes a non-uniform model performance across motion categories, validating the need to move beyond aggregated metrics. For instance, MoMask++ significantly outperforms MDM in FID within categories characterized by subtle motions like \emph{Communication}, \emph{Education}, and \emph{Gestures} (\tabref{tab:category_split_semantic}). To validate these variances, we evaluated 1,000 motion samples \textbf{randomly selected} per category across the top 10 most frequent categories.
These results collectively suggest that diffusion-based approaches (MDM) offer better adherence to physical constraints, while transformer-based approaches (MoMask++) excel at capturing the distributional nuances of complex social and expressive motions.

\subsection{Addressing Physical Quality Discrepancies}

The performance disparities on physical metrics are pronounced (\tabref{tab:category_split_physical}). MoMask++ shows significantly higher values for Jerk (e.g., 332.73 overall vs. 46.77 for MDM) and Acceleration Peaks, indicating less smooth and more artifact-prone sequences. This is typical for discrete tokenization approaches, where the quantization and sequential generation process can lead to accumulated temporal inconsistencies. Conversely, MDM's holistic, non-autoregressive denoising mechanism appears to enforce superior long-range consistency and physical plausibility, leading to substantially smoother motion.

\section{Category-Wise Validation of Superior Motion Dynamics}

Building on the global analysis provided in the main paper, we present a comprehensive per-category comparison of Dynamic Scores between our RoMo dataset and MotionMillion in Figure \ref{fig:comparison_dynamic_score_sup}. This fine-grained breakdown substantiates our primary finding: RoMo consistently exhibits \textbf{higher motion intensity} across the vast majority of categories.

While MotionMillion may achieve higher Dynamic Scores in a few isolated categories, our dataset demonstrates a clear systematic advantage, validating that the substantial global increase in mean Dynamic Score ($0.336$ for RoMo vs. $0.222$ for MotionMillion) is driven not by outliers, but by a systematic improvement in motion quality across the taxonomy. This per-category perspective is essential, as it confirms that RoMo provides a more reliable and vigorous training signal across the entire spectrum of human activities.

\section{Optimizing Training Signal via Dynamic Score Ablation}

To quantify the impact of motion quality on generative performance, we conduct an ablation study across five mutually exclusive data partitions defined by our dynamic score ($S_{Dynamic}$). Starting from the complete dataset (\texttt{All}), we define four distinct subsets ($V_{A}, V_{B}, V_{C}, V_{D}$) based on their dynamic range, testing MDM performance on each.

\subsection{Dynamic Score Partitions}

Starting from the full dataset (\texttt{All}), we define four subsets by increasing the lower-bound threshold for inclusion: $V_A$ ($S_{Dynamic} \ge 0.05$), $V_B$ ($S_{Dynamic} \ge 0.1$), $V_C$ ($S_{Dynamic} \ge 0.15$), and the highly-dynamic set $V_D$ ($S_{Dynamic} \ge 0.5$).

\subsection{Semantic and Physical Insights}

Results in \tabref{tab:table_split_semantic} demonstrate that training on subset $\mathbf{V_C}$ yields the optimal semantic trade-off, achieving the best FID ($17.97$) compared to the full dataset (\texttt{All}, $20.63$). 
This suggests that aggressively filtering sequences with lower dynamic scores effectively removes static noise that dilutes model performance.

However, training exclusively on the lowest dynamic range tested, $\mathbf{V_{D}}$, results in a degraded FID ($28.72$) and significant physical instability (Jerk $113.35$ in \tabref{tab:table_split_physical}). 
This highlights that training solely on motion below the optimal dynamic threshold degrades motion smoothness and generation fidelity. 
It is important to note that while intermediate filtering (e.g., $\mathbf{V_C}$) improves numerical metrics, highly aggressive filtering inevitably discards subtle, fine-grained interactions that are essential for a truly comprehensive and diverse motion distribution.

\begin{table*}[tb]
\centering
\caption{\textbf{Ablation study on motion dynamics.} Evaluation of MDM performance across different data partitions defined by motion dynamic ($S_{Dynamic}$). The subsets range from high-dynamics ($V_A$) to low-dynamics ($V_D$). Results show that selective training ($V_C$) yields better semantic alignment (FID) than using the complete dataset (All).}
\begin{tabular}{lcccccc}
\toprule
\textbf{Method} & \multicolumn{3}{c}{\textbf{R Precision}} & \textbf{Diversity} $\uparrow$ & \textbf{FID} $\downarrow$ & \textbf{Matching Score} $\uparrow$ \\
 & \textbf{Top1} & \textbf{Top2} & \textbf{Top3} &  &  &  \\
\midrule
All & 0.5893 & 0.7652 & 0.8441 & 27.67 & 20.63 & 12.06 \\
VA & 0.5754 & 0.7564 & 0.8473 & 27.66 & 19.61 & 12.18 \\
VB & 0.5863 & 0.7521 & 0.8414 & 27.82 & 18.70 & 12.24 \\
VC & 0.5822 & 0.7578 & 0.8434 & 27.59 & 17.97 & 12.18 \\
VD & 0.4660 & 0.6400 & 0.7441 & 28.79 & 28.72 & 14.26 \\
\bottomrule
\end{tabular}
\label{tab:table_split_semantic}
\end{table*}

\begin{table*}[tb]
\centering
\caption{\textbf{Impact of motion dynamics on physical quality.} Physical metrics for MDM trained on different dynamic score subsets.}
\begin{tabular}{lcccccc}
\toprule
\textbf{Method} & \textbf{Foot Skating} $\downarrow$ & \textbf{Jerk} $\downarrow$ & \textbf{Ground Penetration} $\downarrow$ & \textbf{Floating} $\uparrow$ & \textbf{Acceleration Peaks} $\uparrow$ & \textbf{Dynamic Score} $\uparrow$ \\
\midrule
All & 1.70e-03 & 43.98 & 0.00e+00 & 0.0167 & 0.8171 & 0.2138 \\
VA & 9.61e-04 & 47.69 & 5.99e-06 & 0.0159 & 0.8859 & 0.2311 \\
VB & 1.57e-03 & 45.13 & 3.09e-05 & 0.0161 & 0.8558 & 0.2120 \\
VC & 1.10e-03 & 46.77 & 0.00e+00 & 0.0146 & 0.8374 & 0.2132 \\
VD & 4.23e-04 & 113.35 & 6.94e-04 & 2.85e-03 & 1.37 & 0.2711 \\
\bottomrule
\end{tabular}
\label{tab:table_split_physical}
\end{table*}

\section{Context-Aware Adaptive Filtering}

To demonstrate the unique utility of our hierarchical taxonomy, we present an adaptive filtering approach that takes advantage of category-specific semantic context to substantially improve the quality of the dataset while preserving critical motion diversity. We focus on the \textbf{foot skating ratio} metric—a measure of foot sliding artifacts common in motion capture data.

Traditional \textbf{global filtering} removes all samples with high foot skating values across the entire dataset. 
For instance, removing the top $15\%$ of foot skating ratio samples results in $123,178$ filtered samples. 
However, this indiscriminate approach fails to recognize that high foot skating values represent \textbf{authentic motion characteristics} in categories such as skateboarding, skiing, and snowboarding, while indicating quality issues in sedentary activities like office work or crafting.

Our adaptive filtering approach addresses this semantic ambiguity through the following steps:
\begin{enumerate}
    \item \textbf{Category Identification:} We identify specific "expected skating" categories where foot skating is natural.
    \item \textbf{Authenticity Preservation:} We preserve $100\%$ of samples in these natural categories, regardless of their foot skating values.
    \item \textbf{Contextual Filtering:} We apply \textbf{per-subcategory $15\%$ filtering} to all other categories, removing only the worst samples relative to their specific contextual distribution.
\end{enumerate}

This taxonomy-enabled methodology preserves $\mathbf{62,733}$ high-quality authentic samples that would have been incorrectly removed by global filtering, while simultaneously identifying $\mathbf{45,385}$ context-specific quality issues that global filtering missed. This result demonstrates how our taxonomy provides the essential semantic structure needed to distinguish between authentic motion and quality artifacts, enabling a robust balance between quality improvement and diversity preservation.

The same adaptive filtering paradigm can be universally applied to other evaluation metrics:
\begin{itemize}
    \item \textbf{Floating (vertical displacement):} Expected in jumping and acrobatic categories but signals errors elsewhere.
    \item \textbf{Ground Penetration:} Natural for digging and excavation activities but represents artifacts in most other contexts.
\end{itemize}

This flexibility demonstrates how the taxonomy enables researchers to design custom quality control pipelines tailored to downstream tasks. We leave the exploration and evaluation of advanced adaptive filtering strategies within text-to-motion generation for future work.

\section{Implementation Details}
In the main paper we demonstrated the performance of several SOTA architectures (diffusion and GPT) when trained on our dataset. For reproducibility, we provide the implementation details for the different models.

\textbf{MDM Training Parameters}
We train a Motion Diffusion Model using a transformer-based denoising architecture with 8 layers, 4 attention heads, latent dimension 512, and feedforward dimension 1024. Text conditioning is provided through a frozen BERT encoder with maximum sequence length of 128 tokens. The diffusion process employs 50 denoising steps during training. Motion sequences are processed with a maximum length of 224 frames at 30 FPS, corresponding to approximately 7.5 seconds of motion. We train with batch size 256 for 750,000 iterations using exponential moving average (EMA) of model weights without weight decay regularization. 

\textbf{MoMask++ Training Parameters}
We train MoMask++ on our ITW dataset, with adaptations for our hardware and dataset characteristics. In Stage 1, we train a hierarchical residual VQ-VAE with 6 quantizers across 4 scales (8, 4, 2, 1) using a codebook size of 512 and code dimension of 512. The encoder-decoder architecture uses a width of 512, depth of 3, and temporal downsampling factor of 2. We train for 1000 epochs with batch size 512, learning rate 3e-4, and commitment loss weight 0.02. Critically, we disable gradient clipping to prevent codebook collapse, following recent findings in motion tokenization. In Stage 2, we train a bidirectional masked transformer with 8 layers, 6 attention heads, latent dimension 384, and feedforward dimension 1024. Text conditioning is provided via a frozen T5-base encoder (768-dim embeddings). We train for 500 epochs with batch size 256, applying linear learning rate scaling to 8e-4 (4× the base rate of 2e-4). Both stages use classifier-free guidance with 10\% conditioning dropout and a 2000-step warmup schedule with multi-step learning rate decay at milestones [100k, 150k] with gamma 0.3. All experiments are conducted on a single NVIDIA A100-80GB GPU.

%%%%%%%%%%%%%%%%%%%%%%%%%%%%%%%%%
%%%%%%%%%%% Reference %%%%%%%%%%%
%%%%%%%%%%%%%%%%%%%%%%%%%%%%%%%%%
{
    \small
    \bibliographystyle{ieeenat_fullname}
    \bibliography{main}
}